\documentclass[a4paper,fleqn]{cas-sc}

\usepackage[authoryear,longnamesfirst]{natbib}
\usepackage{tipa}
\usepackage{placeins}
\usepackage{longtable}
\usepackage{booktabs}
\usepackage{multicol}
\usepackage{array}
\usepackage{multirow}
\usepackage{tabularx}
\usepackage{xltabular}
\usepackage[table]{xcolor}
\usepackage{amssymb}
\usepackage{graphicx}
\usepackage{tikz}
\usepackage{float}
\usepackage{pdfpages}
\usetikzlibrary{arrows.meta,decorations.pathreplacing,positioning,calc}

\newcolumntype{Q}{>{\raggedright\arraybackslash}X}

\definecolor{tablegray}{gray}{0.95}
\definecolor{tableline}{gray}{0.75}

\def\tsc#1{\csdef{#1}{\textsc{\lowercase{#1}}\xspace}}
\tsc{WGM}
\tsc{QE}
\begin{document}
\let\WriteBookmarks\relax
\def\floatpagepagefraction{1}
\def\textpagefraction{.001}

\shorttitle{Disentangling Phonological and Morphosyntactic Contributions to ASR Bias}    

\shortauthors{H. Mojarad \& K. Tang}  

\title [mode = title]{Disentangling Phonological and Morphosyntactic Contributions to ASR Bias in African American English}  

\tnotemark[1] 

\tnotetext[1]{} 


%\author[1]{}%[<options>]
\author[1]{Hamid Mojarad}[orcid = 0009-0005-2502-1240]

\cormark[1]

\fnmark[1]

\ead{hamid.mojarad@hhu.de}

\ead[url]{https://www.hamidmojarad.com}

\credit{Conceptualization, Methodology, Software, Validation, Investigation, Data curation, Formal analysis, Visualization, Writing – original draft, Writing – review \& editing, Project administration}
\affiliation[1]{organization={Department of English Language and Linguistics, Institute of English and American Studies, Faculty of Arts and Humanities, Heinrich Heine University Düsseldorf},
            addressline={Universitätsstraße 1}, 
            city={Düsseldorf},
%          citysep={}, % Uncomment if no comma needed between city and postcode
            postcode={40225}, 
            state={North Rhine-Westphalia},
            country={Germany}}

\affiliation[2]{organization={Department of Linguistics, University of Florida},
            addressline={Turlington Hall}, 
            city={Gainesville},
%          citysep={}, % Uncomment if no comma needed between city and postcode
            postcode={32611-5454}, 
            state={Florida},
            country={United States of America}}
            
\author[1,2]{Kevin Tang}[orcid = 0000-0001-7382-9344]

% Footnote of the second author
\fnmark[2]

% Email id of the second author
\ead{kevin.tang@hhu.de}

\ead[url]{https://www.kevintang.org/}

\credit{Conceptualization, Methodology, Formal analysis, Writing – review \& editing, Supervision, Resources, Project administration}

\cortext[1]{Corresponding author}

\fntext[1]{}


\begin{abstract}
Automatic speech recognition (ASR) performs worse for speakers of African American English (AAE), and prior work links this disparity to individual phonological and morphosyntactic features. In natural speech, however, such features co-occur, and it is unknown whether their effects on ASR errors add up or compound. We test this for consonant cluster reduction (CCR) and the null copula in a Corpus of Regional African American Language (CORAAL) sample of 10,036 utterances from 223 speakers, using two hybrid, two self-supervised and one supervised model as released. The null copula had a larger effect on utterance word error rate (WER), whereas CCR mainly affected the exact transcription of the word carrying the cluster and had a relatively small effect on WER, which was significant in only one self-supervised model. When the two features co-occurred, their effects added rather than compounded, although moderate interactions cannot be ruled out. Larger or better-trained variants within a model family lowered overall WER but did not consistently reduce the effect of either feature. These findings show that the consequences of dialectal variation for ASR depend on a feature's linguistic level and on whether it alters a retained word or removes a word from the utterance. They also show that aggregate WER can obscure feature-specific error patterns: evaluating ASR at feature and token levels can reveal errors that it does not, and bias-mitigation training for ASR in linguistically diverse speech requires attention to specific phonological and morphosyntactic variation.
\end{abstract}


\begin{highlights}
\item Consonant cluster reduction and the null copula add rather than compound ASR errors
\item The null copula raises utterance word error rates in ASR systems
\item Consonant cluster reduction mainly harms the word itself, not the whole utterance
\item Larger or better-trained models lower error rates but not individual feature effects
\item  ASR training for bias mitigation should target specific phonological and morphosyntactic variations
\end{highlights}

% Keywords

\maketitle

\section{Introduction}\label{sec:intro}

Automatic speech recognition (ASR) is increasingly built into everyday services, and its integration is expected to provide equitable access for all speakers rather than reinforce existing inequalities \citep{Feng2024, Cunningham_Adjagbodjou_Basoah_Jawara_Kadoma_Lewis_2025}. For speakers of African American English (AAE), this promise is not yet met, and the consequences go beyond wrong transcripts. \citet{Mengesha_2021} report that ASR errors make African Americans feel ``othered'' and leave them feeling that the technology was not made for them, so that they accommodate their speech in order to be recognized. 
AAE older adults likewise envision voice technologies that better understand their cultural context \citep{Brewer_et_al_2023}. Persistent performance disparities \citep{Chen_2025} thus make ASR not only less beneficial but actively harmful, amplifying linguistic discrimination and exclusion \citep{Choi_Choi_2025}, including in hiring and healthcare \citep{Martin2023}.

This problem is not unique to AAE. Dialect effects on ASR accuracy were reported for regional varieties of English early on \citep{Tatman2017, tatman-2017-gender}, and they persist in recent systems: Whisper is more accurate on North American accents than on British and Australian accents \citep{graham2024whisper}, errors on Newcastle English are closely tied to regional dialect features \citep{serditova25_interspeech}, a comparison of 24 varieties of English, most of them non-native, found systematic differences in word and phone error rates \citep{chan22b_interspeech}, and a large-scale audit of popular English ASR services found that performance is related to the geopolitical standing of speakers' countries of birth, even after controlling for linguistic covariates \citep{DiChristofano_2023}. These findings matter because ASR output is increasingly used as a first-draft transcript, for example of conversational speech \citep{Russell2024}, so any disparity in the system is inherited by downstream users.

A common explanation for such disparities is skewed training data \citep{MarklMcNulty2022}. Data certainly matter: AAE is underrepresented in major speech corpora \citep{dorn-2019-dialect}, and it makes up only 0.007--0.18\% of documents in open pretraining corpora and is removed more often than Mainstream American English (MAE) by common quality filters \citep{deas-etal-2025-data}, which is relevant for the language models on which ASR systems rely.

However, data gaps do not act uniformly; they surface through specific linguistic features. \citet{koenecke2020racial} traced the racial disparity primarily to the acoustic models, since the gap was as large on identical phrases. On Newcastle English, phonological variation accounted for most errors, along with local vocabulary and non-standard grammatical forms, and the linguistic features explained errors better than speaker gender or age \citep{serditova25_interspeech}. Systematic differences conditioned by linguistic structure also remain for varieties that a system claims to support \citep{chan22b_interspeech}, and for the low-resource language Irish, dialect-balanced training data still gave unequal performance across dialects \citep{lonergan23_interspeech}. At the same time, phoneme-level analyses of Dutch and Mandarin ASR with hybrid and end-to-end architectures found that pronunciation differences between speaker groups explain only a fraction of the bias, and that mitigation needs language- and architecture-specific solutions \citep{Feng2024}. This suggests a general mechanism: systems fail when the phonological, prosodic, lexical, or morphosyntactic properties of a speaker's variety diverge from the norms encoded in the model, and how much a given gap affects recognition accuracy depends on which features are involved and on how the system uses them.

This question is pressing now because scale has not closed the gap. Racial disparities in commercial ASR were documented several years ago \citep{koenecke2020racial}, and since then systems have grown substantially in training data and model size; Whisper, for example, was trained on 680{,}000 hours of labeled speech \citep{radford2023robust}. Yet a longitudinal evaluation of 16 systems released between 2021 and 2025 on eight datasets of English varieties found that the gap between Mainstream US English and minority accents is not reliably shrinking over time \citep{Metzger2026}, and performance for AAE speakers has been uneven \citep{wassink2022uneven} and remained so for the self-supervised and Whisper models evaluated in 2024 \citep{chang24d_interspeech, Harris2024}. It is therefore an open question whether the remaining disparity will shrink with more data and larger models, or whether it is tied to specific linguistic features and the way these features combine in natural speech. We address the second part of this question: we test how two AAE features, alone and together, affect systems of different architectures and, within two of them, different sizes, so that the feature effects can be compared across the scales we have.

\subsection{Problem statement}\label{sec:problem_statement}

AAE is a rule-governed dialect with distinct phonological and morphosyntactic configurations that differentiate it from MAE \citep{labov1972, Green_2002, rickford1999}. It has long been stigmatized in educational and social contexts, a pattern tied to anti-Black racism and the devaluation of African American children's linguistic practices \citep{Washington2025}. Because its variation is systematic rather than random, AAE is a crucial test case for equity in ASR: its features can be misread or under-accommodated instead of being treated as legitimate structural properties of a speech community \citep{wassink2022uneven}. \citet{koenecke2020racial} evaluated five commercial systems on more than 2{,}000 matched utterances and found roughly double the Word Error Rate (WER) for AAE speakers despite identical lexical content.

Prior work has linked these errors to specific features. Phonologically, \citet{wassink2022uneven} related the errors of one ASR system to a set of sociophonetic variables, among them consonant cluster reduction (e.g., \emph{test} $\rightarrow$ \emph{tes'}), th-stopping, and \emph{-ing} alternation, with consonant cluster reduction the most frequent source of AAE-related errors, and \citet{mojarad25_interspeech} found a small but significant WER increase for consonant cluster reduction and \emph{-ing} reduction. Morphosyntactically, \citet{martin20_interspeech} showed that Google Cloud Speech and Mozilla DeepSpeech make more errors on habitual \emph{be} and its surrounding words than on non-habitual \emph{be}, and \citet{Martin2022thesis} found that common AAE morphosyntactic features (invariant \emph{be}, perfective \emph{done}, and \emph{ain't}) occur at low frequencies in corpora  that are commonly used to train or evaluate ASR systems such as Fisher, Switchboard, LibriSpeech, and TIMIT. Both feature types are therefore sites of failure.

Studies show that utterances with more AAE features have higher error rates \citep{koenecke2020racial, chang24d_interspeech}, and dialect density, the frequency with which dialect features occur in speech, has been formalized and estimated automatically for AAE, including children's speech \citep{johnson22_interspeech, johnson2024dialectdensity}. Such measures, however, pool all features into a single score and therefore do not show whether particular combinations of feature types have a larger effect than the sum of their parts. Most feature-level studies likewise examine one feature at a time, or several features of one type. Intersections of speaker attributes such as gender and dialect have been examined \citep{tatman-2017-gender, Harris2024}, but, to our knowledge, not the intersection of linguistic features themselves, probably because this requires utterances matched on, and annotated for, the presence of each feature, whereas standard evaluations compare speaker groups.

There are reasons to expect that co-occurring features do not simply add up. ASR decoding is sequential and context-dependent, so an error on one word can propagate to its neighbours, a ``domino effect'' \citep{martin20_interspeech}. 
\citet{martin20_interspeech} attribute the stronger influence of preceding words to language models that consider only preceding context. 
For instance, \emph{be comin'} (habitual \emph{be} plus \emph{-ing} reduction) may be misrecognized as \emph{became}, altering tense and aspect \citep{martin20_interspeech}.
An utterance containing both a phonological and a morphosyntactic feature could therefore be harder than the sum of its parts if the first error removes context needed to recover the second (or, in systems that also use following context, the reverse). Acoustic and language-model components both contribute to AAE bias \citep{koenecke2020racial, martin20_interspeech}, so such an interaction would arise at the interface between acoustic evidence and linguistic context, and the direction of the context matters. Moreover, systems differ in how they use it: hybrid pipelines combine an acoustic model with an explicit language model (typically $n$-gram, possibly rescored with neural models in both directions), self-supervised CTC encoders attend to context in both directions and are often decoded without any language model, and end-to-end supervised encoder-decoder models generate text autoregressively. This motivates testing whether the co-occurrence of such linguistic features compounds errors across systems with different architectures, or whether the two effects simply add up.

This gap has two practical consequences. First, aggregate WER can obscure predictive bias \citep{MarklMcNulty2022}, and when multiple linguistic features co-occur within an utterance, it cannot reveal how much each feature, or their interaction, contributes to recognition errors. If co-occurring features compound one another, evaluations that consider features in isolation would underestimate the cumulative effect of natural AAE speech on ASR errors. Second, knowing which combinations have the largest effects on errors matters for mitigation. Rather than fine-tuning ASR systems with dialectal data indiscriminately, mitigation efforts could prioritize utterances containing combinations of features associated with larger effects on errors, whereas existing fairness-oriented fine-tuning has primarily operated at the level of accent groups \citep{swain2025fairasr}. Addressing this gap is therefore essential for developing more equitable ASR technologies.

\subsection{Research question}\label{sec:research_questions}

We operationalize the phonological feature as consonant cluster reduction (CCR) and the morphosyntactic feature as null copula since both features are prevalent in AAE \citep{Green_2013}. We ask the research question with two hypotheses: Does the co-occurrence of a phonological (CCR) and a morphosyntactic (null copula) AAE feature increase ASR errors beyond the sum of their individual effects? 

\begin{itemize}
    \item \textbf{$H_0$:} The co-occurrence of phonological and morphosyntactic features does not produce an interaction effect on ASR error beyond the sum of their individual effects (additive).
    \item \textbf{$H_1$:} The co-occurrence produces an interaction effect that exceeds this sum (multiplicative). 
\end{itemize}

The study is designed to distinguish between these outcomes, not merely to test whether the features are associated with bias. We find that, across five systems (two traditional hybrid models, two self-supervised models, and one supervised model), both features are associated with higher errors, but their effects add up: the combined effect is not larger than the sum of the two separate effects.

Section~\ref{sec:AAE_Features} describes the phonological and morphosyntactic features of AAE, 
Section~\ref{sec:Methodology} presents the data, models and statistical analysis, Section~\ref{sec:Results} the results, Section~\ref{sec:Discussion} their interpretation and limitations, and Section~\ref{sec:Conclusion} concludes.

\section{The speech of African American English} \label{sec:AAE_Features}

African American English (AAE) is a systematically structured variety of English spoken by African Americans across a wide range of regions in the United States, including urban centers such as Atlanta, Philadelphia, and New York City, metropolitan California, and the Mississippi Delta \citep{Kretzschmar_2015, Labov_etal_2015, Blake_etal_2015, Rickford_2015, Wilkerson_2015}. It emerged from the language-contact situation of enslaved Africans and their descendants, who were exposed to British-based English dialects, African languages, and, in some cases, creolized varieties of the African diaspora \citep{Wolfram_2003}. Over generations, these influences merged into a distinct speech community.

AAE shows regional variation alongside a shared core of phonological, morphological, syntactic, semantic, and lexical patterns, making it recognizable as a coherent variety within the broader English-speaking landscape \citep{Wolfram_2015, Green_2002}. It is also a socially and ethnically anchored variety, closely tied to African American identity, community practices, and relations of power \citep{smitherman1977, morgan2002, Washington2025}. This dual status makes AAE especially relevant for equity-oriented research in speech technology: ASR systems trained largely on MAE may treat AAE-specific patterns as noise or error rather than as legitimate variants, thereby reproducing broader sociolinguistic inequalities in educational, legal, and media contexts \citep{Washington2025, heuser24_interspeech}.

AAE has many documented phonological features (e.g., coronal stop deletion, \emph{l}-vocalization, /r/-deletion) and morphosyntactic features (e.g., the null copula, habitual \emph{be}, zero past-tense marking) \citep{Green_2002, rickford1999, Bayley_Villarreal_2019}. This study focuses on one of each: CCR, the simplification of a word-final consonant cluster (\textit{test} $\rightarrow$ \textit{tes}), and the null copula, the omission of present-tense \emph{is} or \emph{are} (\textit{She happy}). These features provide a fine-grained lens on how ASR systems handle AAE-specific variation and on whether phonological and morphosyntactic sources of error add up or compound.

\subsection{Consonant cluster reduction} \label{sec:CCR}

Consonant cluster reduction (CCR) is a systematic phonological process in which word-final consonant clusters, as in \textit{west} or \textit{cold}, are simplified (\textit{wes'}, \textit{col'}) \citep{Wolfram_2017}. Its likelihood is conditioned by the following segment, with reduction more frequent before a consonant than before a vowel, and by cluster type, with monomorphemic clusters such as \textit{wind} reduced more often than clusters formed by the past-tense suffix \textit{-ed}, as in \textit{guessed} \citep{Wolfram_2017}. Both effects were found in English varieties examined by \citet{Wolfram_2017}, although their relative strength differs: the following segment has the larger effect in Standard English and Appalachian Vernacular English, whereas cluster type has the larger effect in Southern white working-class and Southern African American working-class speech. CCR is therefore not specific to AAE: it occurs in  other English varieties with varying rates of reduction.

In AAE, CCR typically involves omission of the final stop in a two-consonant cluster (e.g., \textit{test} \textipa{/tEst/} $\rightarrow$ \textipa{[tEs]}) or of the penultimate consonant in a three-consonant cluster (e.g., \textit{fists} \textipa{/fIsts/} $\rightarrow$ \textipa{[fIs:]}) \citep{Erik_Bailey_2015}. As in other varieties, reduction is more frequent before a consonant than before a vowel and in monomorphemic than bimorphemic clusters \citep{Erik_Bailey_2015, Wolfram_2017}. Additional conditioning factors include unstressed syllables, homorganic clusters (e.g., \textit{cold} $\rightarrow$ \textipa{[kol]}), and the type of following consonant, with reduction especially frequent before obstruents and less frequent before nasals and liquids \citep{Erik_Bailey_2015}. AAE also generally shows higher rates of reduction than other varieties: in the data compared by \citet{Wolfram_2017}, African American working-class speakers reduced 88--97\% of non-\textit{-ed} clusters before a consonant and 72\% before a vowel, compared with 66\% and 12\% for Standard English. Reduction further increases as the first consonant of the cluster rises in sonority \citep{labov1972, wolfram_Thomas2008}.

CCR is relevant to ASR because reduced forms may be misinterpreted as omissions or segmental errors even though they are regular dialect features. A neural ASR study found a small but significant increase in WER attributable to CCR \citep{mojarad25_interspeech}, and \citet{wassink2022uneven} found that CCR was among the sociophonetic variables most strongly associated with ASR errors across speaker groups. Neither study, however, examined how CCR combines with a morphosyntactic feature. Such a combination is plausible because errors on one word can propagate to following words \citep{martin20_interspeech}: if a reduced cluster is misrecognized, the resulting error may reduce the contextual evidence available for recovering an absent copula. Because ASR systems differ in how much preceding and following context they use, CCR and the null copula provide a natural test of whether phonological and morphosyntactic features combine additively or compound ASR errors.

\subsection{Null copula} \label{sec:NullCopula}

The null copula, or zero copula, refers to the systematic omission of overt copular or auxiliary \emph{be} in present-tense contexts where MAE would require it, as in \textit{She \textipa{[}ø\textipa{]} the first one started us off} \citep{labov_1969} or \textit{They \textipa{[}ø\textipa{]} walking too fast} \citep{Green_2002}. In AAE, deletion is restricted to present-tense \emph{is} and \emph{are}, not past-tense \emph{was} or \emph{were}, and occurs only in environments where MAE permits contraction \citep{labov_1969}. Because of these restrictions, the null copula is a reliable sociolinguistic marker of AAE syntax \citep{labov1972, CraigWashington2006}.

For ASR, the absence of an overt copular form may create a mismatch between the expected lexical realization (\emph{is} or \emph{are}) and the acoustic signal, potentially leading to substitution, deletion, or boundary errors when a system fails to recover the underlying grammatical structure. Earlier work on other AAE morphosyntactic features points in this direction \citep{martin20_interspeech, Martin2023}, but we are not aware of an ASR study of the null copula itself. By treating the null copula as a systematic AAE-specific morphosyntactic feature rather than as noise, this study measures its effect on ASR systems of different architectures and tests how it combines with CCR.

\section{Materials and methods}\label{sec:Methodology}
\subsection{Data Collection}\label{sec:Corpus}
\paragraph{The corpus.}
The Corpus of Regional African American Language (CORAAL) \citep{farrington_corpus_2021} is our data source. It contains
sociolinguistic interviews with African American speakers, with time-aligned orthographic transcriptions in TextGrid format at the utterance and word/phone levels. The complete corpus (version 2023.06, which added the Detroit component) has eight components recorded in different locales and periods, with 271 audio files from 222 primary speakers and about 159.5 hours of speech. Speakers are stratified by gender, age and, for some components, socioeconomic class; age groups and class categories are defined separately for each component (class is coded only for DCA, DCB and DTA), and we describe how we handled them in Section~\ref{sec:StatisticalAnalysis}.

\paragraph{Our selection.}
We use all eight components, but retained only interviewee speech, and only the utterances that contain a site of the
phonological feature (CCR), a site of the morphosyntactic feature (copula), or both (Section~\ref{sec:DatasetDesign}). The resulting utterance-level dataset has 10,036 utterances from 223 speakers (113 male and 110 female) in 268 of the 271 recordings. Table~\ref{tab:corpus_speakers} gives the components and our selection from each.

The multi-dimensional stratification of these speakers ensures diversity in terms of geography, gender, age, and, where it is coded, social class, supporting the analysis of AAE‑specific features and enabling more robust generalization of ASR‑related interaction and bias effects across speaker subgroups.

\begin{table}[h]
\centering
\footnotesize
\caption{CORAAL components and our selection from each (interviewee speech only). The counts are for our selection.}
\label{tab:corpus_speakers}
\begin{tabular}{lllrrr}
\toprule
\textbf{Component} & \textbf{Location} & \textbf{Recorded} & \textbf{Speakers (m / f)} & \textbf{Utterances} & \textbf{Class coded} \\
\midrule
ATL & Atlanta, GA          & 2017--2018 & 13 (8 / 5)   & 880   & no  \\
DCA & Washington, DC       & 1968--1969 & 68 (40 / 28) & 2,390 & yes \\
DCB & Washington, DC       & 2015--2017 & 50 (24 / 26) & 2,277 & yes \\
DTA & Detroit, MI          & 1966       & 40 (18 / 22) & 1,391 & yes \\
LES & Lower East Side, NYC & 2008--2009 & 10 (5 / 5)   & 530   & no  \\
PRV & Princeville, NC      & 2003--2004 & 16 (7 / 9)   & 1,034 & no  \\
ROC & Rochester, NY        & 2016--2017 & 14 (5 / 9)   & 759   & no  \\
VLD & Valdosta, GA         & 2017--2019 & 12 (6 / 6)   & 775   & no  \\
\midrule
Total & & & 223 (113 / 110) & 10,036 & \\
\bottomrule
\end{tabular}
\end{table}

\subsection{Feature annotation and extraction}\label{sec:FeatureExtraction}
We operationalize AAE‑specific features at both the phonological and morphosyntactic levels, using a combination of automated alignment and manual annotation.

\paragraph{Phonological features.}
In this study, we focus exclusively on two-consonant clusters (e.g., \textit{test}, \textit{hand}) for CCR and leave three-consonant clusters (e.g., \textit{texts}) for future work. To extract CCR, we adopted a forced‑alignment pipeline inspired by \cite{Kendall_ing_2021}, who demonstrated that machine‑learning and forced‑alignment approaches can approximate human coding of the sociolinguistic variable (ING-reduction). Critically, CCR was not manually annotated in our dataset; instead, we rely on the automated alignment process as a proxy for phonological variation. 
This is necessary because CORAAL transcripts follow standard‑like orthography and do not represent phonological variants such as CCR: when it is unclear whether a form reflects a phonological process or a morphosyntactic difference, transcribers were instructed to use the standard spelling, so the canonical or reduced realization cannot be read from the transcript. \citet{Kendall_ing_2021} show that automated coding based on pronunciation expansion and forced alignment yields performance close to that of human coders, whose inter-analyst agreement was approximately 88\%, supporting the use of such methods for systematic phonological feature extraction.

Building on this, we use the Montreal Forced Aligner (MFA, version 2.2.17) \citep{mcauliffe17_interspeech} together with the Carnegie Mellon University (CMU) Pronouncing Dictionary, extended to accommodate AAE‑related variability. Words missing from the CMU dictionary were first identified, after which we trained a grapheme‑to‑phoneme (G2P) model on the CMU lexicon and generated pronunciations for these items. The resulting pronunciations were manually inspected and added to the dictionary. For words susceptible to CCR, we added reduced pronunciation variants to the dictionary, derived from every canonical CMU pronunciation of the word (e.g., \textit{accent} \texttt{AH0 K S EH1 N T} $\rightarrow$ \texttt{AH0 K S EH1 N}). In this context, we use the term \textit{canonical} to refer to the full, standard lexical pronunciation (including all consonants in CCR), and \textit{reduced} to refer to the phonologically simplified variant in which the final consonant of the cluster is deleted. This explicit distinction between canonical and reduced forms underlies our subsequent binary coding of CCR at the token level (Section ~\ref{sec:StatisticalAnalysis}).

Using MFA's \texttt{train} command, we trained a custom acoustic model on all eight CORAAL components. The complete corpus was then force-aligned using this trained model with the expanded CMU pronunciation dictionary, and the resulting alignments formed the basis for classifying words as canonical or reduced throughout the study. 

Our reliance on MFA for phonological feature extraction rests on two properties: accurate boundary placement and accurate selection among pronunciation variants listed in the dictionary. Regarding the former, MFA still outperforms modern end-to-end ASR-based aligners on word- and phone-level alignment~\citep{rousso24_interspeech}; and MFA 3.0 reaches state-of-the-art or near state-of-the-art accuracy on English, Japanese and Korean, with mean boundary errors below 15\,ms~\citep{mcauliffe26_interspeech}. Regarding the latter, studies that supply the aligner with competing surface variants and compare its choices with manual coding report agreement comparable to that between human annotators: for English (ING) with MFA~\citep{Kendall_ing_2021} and with the Penn aligner~\citep{yuan11_asru}, for (th)-fronting, (td)-deletion and (h)-dropping in British English~\citep{bailey2016}. Together, these results support MFA as a reasonable basis for extracting phonological features under dialectal variation, provided that the dictionary encodes the relevant dialectal variants.

To estimate this reliability on CORAAL, we compared MFA's selection between canonical and reduced CCR variants with manual coding by two annotators on a selection of CORAAL tokens. On the 2,714 CCR tokens coded by both annotators, MFA's agreement with one annotator (64.0\%, $\kappa=0.29$) was comparable to the agreement between the annotators (60.3\%, $\kappa=0.18$), whereas its agreement with the other was much lower (41.1\%, $\kappa=0.06$). MFA's variant choices are therefore about as consistent with manual coding as manual coding is with itself, but this holds for only one of our two annotators. The generally low $\kappa$ values suggest that the reduced CCR token is difficult to code, plausibly because it is a gradient phenomenon conditioned by several interacting factors~\citep{mojarad26_interspeech}, unlike the more categorical (ING) variation studied by \citet{Kendall_ing_2021}.

\paragraph{Morphosyntactic features.}
For null copula, we rely on manual annotation of AAE-specific utterances as well as a fine-tuned BERT model trained on the annotated null copula data.

In contrast to phonological variants, morphosyntactic variants such as copula absence are transcribed as produced in CORAAL, so they can be identified from the transcripts through annotation. In the DCA and DTA subcorpora, we hand-coded instances of copula and auxiliary absence following established sociolinguistic protocols~\citep{labov_1969, Green_2002}, marking each potential site of \emph{be} deletion and classifying it as present or absent. In the DCB subcorpus, we use the manual annotation of copula and auxiliary absence provided by the CORAAL team, specifically the hand-coded dataset ``R (cop/aux) modeling examples'' (2,500 tokens) and associated R code prepared  by \citet{Kendall_chapter_forthcoming}.
This dataset includes carefully coded copula and auxiliary tokens from the CORAAL Washington, DC (DCB) collection, which we map to our own feature-coding schema to ensure consistency across subcorpora.

Using the annotated null copula dataset, we fine-tuned a BERT-based three-class classifier to distinguish copula presence, copula absence, and irrelevant instances. The model was trained for up to five epochs using a class-weighted loss function and early stopping based on validation performance. The training set contained 4,545 instances (1,832 copula, 880 no-copula, and 1,833 irrelevant instances). The validation and test sets each contained 300 instances, balanced across the three classes (100 instances per class).

On the held-out test set, the model achieved 99.67\% accuracy and macro-averaged precision, recall, and F1-score of 99.67\%. The class-wise F1-scores were 99.50\% for copula presence, 99.50\% for copula absence, and 100\% for irrelevant instances. The model made only one error among the 300 test instances, misclassifying a copula instance as no copula. These results indicate strong performance on the held-out annotated data. We then used the model to identify null copula in the remaining subcorpora (ATL, LES, PRV, ROC, and VLD) and manually double-checked its output. The resulting annotation can therefore be considered manually verified for this feature.

To include overt copulas, we used two sources. For the Interaction dataset (Section~\ref{sec:DatasetDesign}), we took the overt-copula utterances mostly from the hand-coded list released by the CORAAL team (see above), which comes from the DCB subcorpus. For the Copula-only dataset (Section~\ref{sec:DatasetDesign}), we searched the other subcorpora with our fine-tuned BERT tagger. We did not use the utterances that were already annotated as null copula (and used to train the tagger), so that the tagger could only select new utterances. The tagger's output was then checked manually, and we checked all of the selected utterances. Each copula instance was also coded as a main-verb copula (e.g., \textit{she [ø] happy}) or an \textit{-ing} progressive auxiliary (e.g., \textit{they [ø] walking}).

Taken together, the phonological pipeline and the morphosyntactic annotation yield a feature-rich representation of AAE that underlies our subsequent analysis of the effects of CCR and null copula, and of their interaction, in ASR outputs.

\subsection{Dataset design}\label{sec:DatasetDesign}

To design datasets that appropriately address our research question, we first need to consider how each feature is manifested and how it can be reflected in ASR performance. CCR is realised on a specific word that is present in the transcript. The reduced token can therefore be located, and its effect on ASR performance can be assessed both on the token itself, by whether it was transcribed exactly, and on the utterance containing it, through the utterance WER. The null copula, in contrast, is manifested as the absence of a copula somewhere in the utterance. Since there is no word to examine at the position of the copula, its effect can only be assessed through the utterance WER. These differences in how the two features are realised motivate our use of two levels of datasets.

At the \emph{utterance level}, each row represents one utterance, and the outcome is its WER. This level allows us to examine the overall effect of both features on utterance-level ASR performance and is the only level at which the effect of the null copula can be assessed.

However, an utterance-level representation does not fully capture the characteristics of CCR. CCR is very frequent in AAE, and consequently, many utterances contain more than one CCR token. Representing each utterance only once does not capture how many CCR tokens it contains or which specific tokens were misrecognised.

To address this, we created an additional \emph{token-level} version of the dataset, in which each row represents one CCR token. This was achieved by expanding each utterance into as many rows as it contains CCR tokens. For example, an utterance containing three CCR tokens is represented once in the utterance-level dataset but three times in the token-level dataset, with one row corresponding to each CCR token. This representation allows us to examine the transcription accuracy of individual CCR tokens, while retaining the WER of the utterance to which each token belongs.

\paragraph{What we mean by an utterance.}
We use the term \textit{utterance} as CORAAL defines it. In CORAAL, an utterance is a stretch of uninterrupted speech by one speaker, delimited at pauses (the transcribers' criterion is a pause of 60--70 ms; \citealp{farrington_corpus_2021}).
An utterance is therefore not necessarily a complete sentence: it can be a full sentence, part of one, or an
incomplete phrase, and it can also span more than one sentence. What matters for our design is that a feature is assigned to the utterance in which it occurs, and not to the surrounding utterances that may belong to the same sentence. Consequently, two features co-occur only if they occur within the same utterance, between the same two pauses. In the example in Table~\ref{tab:utterance_example}, the sentence \textit{They almost alike, except you have to be in a line of some sort} is split by a pause into two utterances, and the second utterance continues into a further sentence (\textit{And then she call your name\ldots}). The null copula (\textit{They} [are] \textit{almost alike}) and a CCR token (\textit{almost}, a two-consonant cluster) both fall in the first utterance, so they co-occur. The second utterance contains CCR tokens (e.g., \textit{except}, \textit{and}, \textit{don't}) but no null copula, so the two features do not co-occur there.

\begin{table}[ht]
\centering
\footnotesize
\caption{Example of CORAAL utterances (file DCA\_se1\_ag1\_f\_02\_01). The sentence is split at a pause into two utterances, and the second utterance spans two sentences. The null copula and a CCR token occur in the first utterance.}
\label{tab:utterance_example}
\begin{tabularx}{\linewidth}{r l >{\raggedright\arraybackslash}X r r}
\toprule
Line & Speaker & Content & Start (s) & End (s) \\
\midrule
272 & \texttt{DCA\_se1\_ag1\_f\_02} & They almost alike, & 291.01 & 291.99 \\
273 & \texttt{DCA\_se1\_ag1\_f\_02} & (pause 0.41) & 291.99 & 292.40 \\
274 & \texttt{DCA\_se1\_ag1\_f\_02} & except you have to be in a line of some sort. And then she call your name, but Simon Say you don't call your name. & 292.40 & 297.31 \\
275 & \texttt{DCA\_se1\_ag1\_f\_02} & (pause 0.41) & 297.31 & 297.71 \\
\bottomrule
\end{tabularx}
\end{table}

\paragraph{Dataset composition and representation} 
Our research focuses on utterances in which phonological and morphosyntactic features co-occur. Accordingly, the \textbf{Interaction} dataset forms the core of our experimental design. However, to gain a better understanding of the effects of their co-occurrence, it is first necessary to examine the contribution of each feature independently, in utterances where it occurs without the presence of the other feature. Having established the individual effects of the two features, we can then turn to their interaction in utterances where they co-occur, investigating how their combined presence affects ASR performance. Therefore, we use three datasets, summarised in Table~\ref{tab:datasets}. The \emph{\textbf{Interaction}} dataset contains utterances with a copula instance and at least one CCR token. The \emph{\textbf{CCR-only}} dataset contains utterances with CCR tokens and no copula instance, and allows us to estimate the effect of CCR in the absence of a copula instance. The \emph{\textbf{Copula-only}} dataset contains utterances with a copula instance and no CCR token, and allows us to estimate the effect of null copula in the absence of CCR.

These three datasets form mutually exclusive groups and together account for all 10,036 utterances. Every dataset is evaluated on the same ASR systems (Section~\ref{sec:Models}), so comparisons across features use the same systems.

\begin{table}[t]
\centering
\footnotesize
\caption{Dataset composition, units of representation and feature counts.}
\label{tab:datasets}
\setlength{\tabcolsep}{5pt}
\begin{tabular}{@{}l l r r r r@{}}
\hline
 & & & & \textbf{Copula} & \textbf{CCR} \\
\textbf{Dataset} & \textbf{Level (one row per)} & \textbf{Rows} & \textbf{Speakers} & overt / null & reduced / non-reduced \\
\hline
CCR-only    & Utterance  & 4,891 & 173 & --            & 3,886 / 1,005 \\
            & CCR token  & 7,053 & 173 & --            & 5,121 / 1,932 \\
\hline
Copula-only & Utterance  & 2,706 & 195 & 1,575 / 1,131 & --            \\
\hline
Interaction & Utterance  & 2,439 & 196 & 1,034 / 1,405 & 2,065 / 374   \\
            & CCR token  & 4,252 & 196 & 1,833 / 2,418 & 3,300 / 951   \\
\hline
\end{tabular}
\end{table}

At the utterance level, each feature is coded as present or absent in the utterance. An utterance is coded as \textit{reduced} for CCR if it contains at least one reduced CCR token, and as \textit{canonical} otherwise. It is coded as \textit{null copula} if it contains an absent or deleted copula, and as \textit{overt copula} if it contains a present copula. Copula instances are further classified as main-verb copula or \textit{-ing} progressive auxiliary, which allows us to analyse the two environments separately.

For the token-level representation, we expand the CCR-only and Interaction datasets to include one row per CCR token. These datasets are therefore represented at both the utterance and token levels. The copula condition remains a property of the utterance and is therefore assigned to each CCR token belonging to that utterance. The Copula-only dataset, in contrast, is represented at the utterance level only, since it contains no CCR tokens and therefore needs no expansion (Table~\ref{tab:datasets}).

The target utterances inevitably contain other AAE features, such as person-number disagreement, double negation, and \textit{l}-vocalization. Only CCR and null copula were annotated (Section~\ref{sec:FeatureExtraction}), and the analyses are restricted to them.

\subsection{ASR models}\label{sec:Models}
\subsubsection{Model selection and design rationale}\label{sec:ModelSelection}

We compare five open-source ASR systems from three architecture families that differ in acoustic modelling, language modelling, and training data: a hybrid pipeline with separate acoustic and language models (Kaldi; \citealp{Povey_ASRU2011}), a self-supervised encoder fine-tuned for CTC decoding (wav2vec 2.0; \citealp{baevski2020wav2vec2}), and a supervised end-to-end encoder--decoder model (Whisper; \citealp{radford2023robust}). Including all three families allows us to ask whether AAE-related transcription effects persist across architectures. Where possible, we also include two variants of the same architecture to separate model size or training data from architecture: two Kaldi systems trained on different amounts of data, and two wav2vec 2.0 models of different capacity that share their training data. Whisper is represented by a single size.

All systems were used exactly as released, without adaptation or fine-tuning, so the effects we measure are those of publicly available systems. We characterize each system by its \textit{model profile}: architecture, capacity, training data, and decoding.

\paragraph{Kaldi GigaSpeech S and L.}
The Kaldi systems use a hybrid chain architecture in which the acoustic model, lexicon, language models, and decoding graph remain separate. A GMM--HMM system provides alignments; audio is augmented in volume and speed, and i-vectors are appended to the acoustic features. The neural acoustic model combines convolutional, factorized TDNN (TDNN-F), attention, and LSTMP layers, trained with cross-entropy and lattice-free MMI \citep{chen21o_interspeech}. Decoding uses a 4-gram language model followed by forward and backward RNNLM rescoring.

Our first Kaldi choice was the standard LibriSpeech chain recipe. It would have been a natural counterpart to wav2vec 2.0, because both rest on roughly 1,000 hours of read speech (960 hours of LibriSpeech). However, its acoustic model was trained on read audiobook speech and its language model on LibriSpeech transcriptions, which are far from the conversational AAE speech in our data; in pilot runs, this mismatch was too large for its output to be usable. We therefore used the Kaldi baseline for GigaSpeech \citep{chen21o_interspeech}, a multi-domain English corpus combining read and spontaneous speech from audiobooks, podcasts, and YouTube. We use two systems trained on different subsets of the corpus: GigaSpeech S (250 hours) and GigaSpeech L (2,500 hours). The two systems share a recipe and differ primarily in training-data size.

\paragraph{wav2vec 2.0 base and large.}
The wav2vec 2.0 systems use a convolutional feature encoder and Transformer context network trained with a self-supervised contrastive objective, followed by a CTC output layer that maps representations directly to character sequences. The base model has 12 Transformer blocks and about 95M parameters; the large model has 24 blocks and about 317M parameters \citep{baevski2020wav2vec2}. Both checkpoints were fine-tuned for CTC on 960 hours of LibriSpeech and are decoded greedily, without an external language model or post-processing. Because the two checkpoints share their training data and recipe, their comparison isolates model capacity.

\paragraph{\texttt{Whisper tiny.en}.}
Whisper is an encoder-decoder Transformer trained on 680k hours of weakly supervised audio-transcript pairs collected from the internet. It uses log-mel features and generates punctuation and capitalization autoregressively, combining acoustic and language modelling in a single sequence-generation model. We use \texttt{openai/whisper-tiny.en}, the smallest English-only checkpoint \citep{radford2023robust}, with 39M parameters. We include only this size because it provides a low-error, generative counterpart to the other systems; adding larger checkpoints would not broaden the range of architectures or decoding strategies examined.

Taken together, the five systems vary systematically in architecture, capacity, training data, and decoding: a hybrid pipeline with explicit acoustic and language models trained on 250 or 2,500 hours of multi-domain English; a self-supervised CTC encoder at two capacities trained on 960 hours of read speech and decoded without a language model; and a small generative supervised model trained on 680k hours of weakly supervised data. This design allows us to examine whether the effects of CCR, the null copula, and their co-occurrence depend on architecture, capacity, or training data in spontaneous AAE speech.

\subsubsection{Transcription pre- and post-processing}\label{sec:TranscriptionProcessing}

\paragraph{Chunking} To ensure comparability across \textit{Kaldi}, \textit{wav2vec 2.0}, and \textit{Whisper}, all audio was resampled to 16~kHz and segmented into chunks of at least 30 seconds. Each chunk was centered on a target utterance and extended to include at least five complete preceding and five complete following utterances from the same speaker, yielding at least 11 complete utterances per chunk. Chunk boundaries were adjusted to avoid cutting utterances; when necessary, additional preceding utterances were added to reach the minimum duration. This provided each ASR system with the target utterance embedded in sufficient acoustic and conversational context, and each system transcribed these chunks.

\paragraph{Normalization and alignment} ASR outputs and gold-standard references were normalized by removing non-speech tags (e.g., \texttt{<laugh>}) and special markers such as \texttt{/unintelligible/} and \texttt{/redactedspeech/}, by removing punctuation, and by lowercasing all text. We then aligned each full reference chunk with its ASR output at the word level using the Needleman-Wunsch global alignment algorithm \citep{NEEDLEMAN1970443}, which captures substitutions, insertions, and deletions. From the chunk-level alignment, we extracted the span corresponding to the target utterance by locating its token sequence in the reference. We computed WER using \texttt{jiwer}~\footnote{\url{https://github.com/jitsi/jiwer}} for the target utterance (Section~\ref{sec:OverallResult}). The target-utterance alignments were subsequently used for linguistic analyses.

\subsection{Statistical analysis}\label{sec:StatisticalAnalysis}

\subsubsection{The experiments}\label{sec:Experiments}
\begin{figure}[!htbp]
\centering
\includegraphics[width=\linewidth]{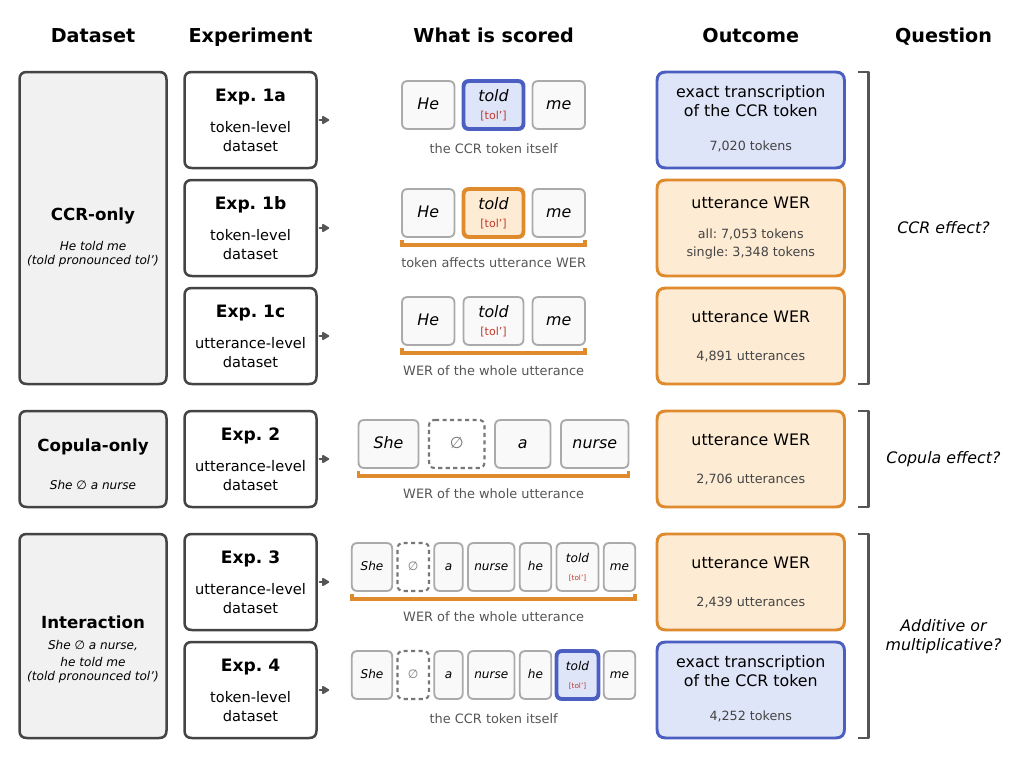}
\caption{Design of the four experiments. Each dataset (left) is defined by the features present in the utterance: CCR (canonical vs.\ reduced token) and the copula (present vs.\ deleted, $\varnothing$). The experiments differ in the unit of analysis (token-level or utterance-level dataset) and therefore in what is scored (middle): the CCR token itself, or the WER of the whole utterance. Reference words are shown in canonical spelling; the reduced pronunciation of a CCR token is marked in red in square brackets. A dashed box is an absent word (deleted copula). In the outcome boxes, blue marks a binary outcome (logistic model) and orange a continuous outcome (linear model), and the numbers give the analysed tokens or utterances (single = tokens from utterances with exactly one CCR token). Experiments~1 and~2 estimate the effect of each feature alone; Experiments~3 and~4 test whether the two effects are additive or multiplicative.}
\label{fig:experiments}
\end{figure}

We investigate whether the effects of CCR and the null copula on ASR errors are additive ($H_0$) or multiplicative ($H_1$) when the two features co-occur within the same utterance. Following the rationale in Section~\ref{sec:DatasetDesign}, we first examine the effect of each feature in utterances where it occurs alone and then test how the two effects combine in utterances where both occur. This is done through four experiments (Figure~\ref{fig:experiments}): Experiment~1 examines CCR alone and Experiment~2 the null copula alone, while Experiments~3 and~4 examine the two features together. Each experiment is conducted separately for each of the five ASR systems, resulting in 20 sets of analyses.

\paragraph{\textbf{Experiment 1: The individual effect of CCR.}}
Experiment~1 examines the effect of CCR on ASR performance in the CCR-only dataset, which contains utterances with CCR but no copula instance. It investigates whether the realization of CCR (reduced vs.\ canonical) is associated with (i) the transcription accuracy of individual CCR tokens (Experiment~1a) and (ii) the WER of the utterances in which they occur (Experiments~1b and~1c).

The two WER analyses use different levels of representation and address complementary questions. Experiment~1b is specific to individual CCR tokens: because a reduced token can be located, it asks whether the realization of an individual CCR token predicts the WER of the utterance in which that token occurs. CCR is very frequent in AAE, and utterances typically contain more than one CCR token. Consider the utterance \textit{They just told me the best idea}, which contains three CCR tokens (\textit{just}, \textit{told}, \textit{best}). In the token-level dataset used for Experiment~1b, it appears three times, once per token, and each row carries the same utterance WER. This isolates the effect of individual tokens on WER, but it cannot capture how many CCR tokens the utterance contains, and the same WER is counted once for each token.

This is exactly where Experiment~1c comes in. It includes each utterance once, with all of its CCR tokens represented at the utterance level rather than expanding the utterance into separate rows for each CCR token. As a result, the utterance WER is not repeated once for every CCR token. An utterance is coded as reduced if at least one of its CCR tokens is reduced, and as canonical otherwise. The reduced/canonical distinction is therefore a property of the utterance rather than of an individual token, unlike in Experiment~1b. To account for the number of CCR tokens that may affect the utterance WER, the total number of CCR tokens is included as a control; in the example above, this value is three. This approach captures the aggregate effect of reduced CCR in the utterance, takes the frequency of CCR into account, and counts each utterance WER value once. It is also the approach used for the null copula: because the null copula is the absence of a word, there is no token whose effect on WER can be examined. It can therefore only be coded as a property of the utterance (present or absent), and its effect is measured on the WER of that utterance. Coding CCR in the same way in Experiment~1c therefore makes it directly and fairly comparable with the null copula.

\paragraph{Experiment~1a: Exact transcription of the token.}
This analysis examines whether the realization of a CCR token predicts whether the ASR system transcribes that token exactly. It uses the token-level CCR-only dataset, with one row per CCR token. The binary outcome, \texttt{Exact}, is coded as 1 if the ASR transcription matches the reference token exactly and 0 otherwise, so the model is a logistic mixed-effects model (\texttt{glmer} in \texttt{lme4}; \citealp{Bates2015}). The predictor of interest is \texttt{CCR\_tok}, the realization of the token according to the forced alignment (MFA): reduced $=-0.5$ and canonical $=+0.5$. 

The following model represents the intended prediction structure. It was tested separately for each ASR system, with nested model comparisons used to determine the final model:
\begin{quote}\small\raggedright
\texttt{Exact \textasciitilde{} CCR\_tok + (1 + CCR\_tok | Speaker) + (1 + CCR\_tok | Word) + (1 | Utterance)}
\end{quote}

\paragraph{Experiment~1b: WER of the utterance (token level).}
This analysis examines whether the realization of an individual CCR token predicts the WER of the utterance in which it occurs. It uses the token-level CCR-only dataset, with each row representing one CCR token and carrying the WER of its utterance. Because an utterance can contain several CCR tokens, the same WER can occur in several rows (repeated WER), which may affect the estimated standard errors. We therefore examine two versions of the dataset: all 7,053 CCR tokens from 4,891 utterances (\textit{all-tokens}), with a random intercept for utterance tested to account for the repeated values, and a \textit{single-token} version with the 3,348 utterances that contain exactly one CCR token, in which each utterance contributes one row and no WER value is repeated, which provides a complementary test.

The intended model structure, tested separately for each ASR system and dataset version, is:
\begin{quote}\small\raggedright
\texttt{WER \textasciitilde{} CCR\_tok + (1 + CCR\_tok | Speaker) + (1 + CCR\_tok | Word) + (1 | Utterance)}
\end{quote}
The utterance random intercept applies only to the all-tokens version.

\paragraph{Experiment~1c: WER of the utterance (utterance level).}
Whereas Experiment~1b examines the realization of individual CCR tokens, this analysis examines the overall realization of CCR within an utterance. It uses the utterance-level CCR-only dataset, with one row per utterance (4,891 utterances containing at least one CCR token), so WER is not repeated.

The predictor of interest, \texttt{CCR\_utt}, indicates whether an utterance contains at least one reduced CCR token (reduced $=-0.5$) or only canonical CCR tokens (canonical $=+0.5$). Thus, unlike Experiment~1b, this analysis does not distinguish between individual tokens or their positions within the utterance; it captures whether reduced CCR is present anywhere in the utterance. The total number of CCR tokens in the utterance, \texttt{CCR\_total}, is included as a control so that the effect of reduced CCR is estimated independently of the overall frequency of CCR in the utterance.

The intended model structure, tested separately for each ASR system, is:
\begin{quote}\small\raggedright
\texttt{WER \textasciitilde{} CCR\_utt + CCR\_total + (1 + CCR\_utt | Speaker)}
\end{quote}

\paragraph{\textbf{Experiment~2: The individual effect of null copula.}}
This analysis examines the effect of null copula on ASR using the utterance-level copula-only dataset. It uses utterances that contain a copula instance (present or deleted) and no CCR instance, with one row and one WER value per utterance. The predictor of interest is \texttt{NC} (null copula), a property of the utterance: copula deleted $=-0.5$ and copula present $=+0.5$. The model is specified as follows:
\begin{quote}\small\raggedright
\texttt{WER \textasciitilde{} NC + (1 + NC | Speaker)}
\end{quote}

The preceding experiments assessed the effects of CCR and null copula separately. Experiments~3 and~4 examine whether their co-occurrence within the same utterance affects ASR performance. We address this question at two levels of analysis. At the utterance level (Experiment~3), we test whether the co-occurrence of CCR and null copula is associated with overall utterance WER. At the token level (Experiment~4), we test whether a null copula elsewhere in the utterance affects the exact transcription of individual CCR tokens. The two analyses examine the effect of the interaction on global utterance accuracy and on the recognition of individual CCR tokens, respectively.

\paragraph{\textbf{Experiment~3: Interaction effect on WER of the utterance.}}
This experiment assesses the effect of CCR and null-copula co-occurrence on utterance-level WER. It uses the utterance-level version of the Interaction dataset. The predictors are \texttt{CCR\_utt}, \texttt{NC}, and, in the interaction model, their product; \texttt{CCR\_total} is again included as a control. \texttt{CCR\_utt} is operationalized exactly as in Experiment~1c: an utterance is coded as reduced if it contains at least one reduced CCR token, regardless of the number or position of such tokens, and as canonical if all its CCR tokens are canonical. We fit the following additive and interaction models separately for each ASR system:
\begin{quote}\small\raggedright
Additive: \texttt{WER \textasciitilde{} CCR\_utt + NC + CCR\_total + (1 + CCR\_utt | Speaker)}\\
Interaction: \texttt{WER \textasciitilde{} CCR\_utt * NC + CCR\_total + (1 + CCR\_utt | Speaker)}
\end{quote}

\paragraph{\textbf{Experiment~4: Interaction effect on exact transcription of the token.}}
This analysis uses the token-level version of the Interaction dataset to examine whether the co-occurrence of CCR and null copula affects the exact transcription of CCR tokens. Specifically, it tests whether the effect of reduced CCR on transcription accuracy differs depending on whether the utterance also contains a null copula. The additive and interaction models for this experiment are specified as follows:
\begin{quote}\small\raggedright
Additive: \texttt{Exact \textasciitilde{} CCR\_tok + NC + (1 + CCR\_tok | Speaker) + (1 + CCR\_tok | Word) + (1 | Utterance)}\\
Interaction: \texttt{Exact \textasciitilde{} CCR\_tok * NC + (same random effects)}
\end{quote}

Experiments~3 and~4 were repeated separately for utterances in which the copula functions as a main verb and in which it functions as the progressive auxiliary (\texttt{ING}~$=0$ and $1$), because the two copula functions may differ in their effect on ASR performance.

The experiments described above specify the dataset, unit of analysis, outcome, and predictors relevant to each analysis. The random-effects structures shown represent the candidate terms considered during the forward selection procedure described in Section~\ref{sec:ModellingProcedure}; only terms retained through this procedure are included in the final models (Appendix~\ref{app:final_models}). All models additionally include \texttt{Age} (four levels) and \texttt{Gender} as demographic control variables, although these are omitted from the formulas above for brevity.

\subsubsection{Coding, model selection and criticism}\label{sec:ModellingProcedure}
All predictors of interest were contrast-coded at $\pm0.5$ (reduced CCR and null copula $=-0.5$; canonical CCR and overt copula $=+0.5$), so that each main effect is the average difference between its two levels and an interaction coefficient is the difference between the two simple effects. With this coding, reduced CCR or a null copula is associated with worse recognition when the coefficient is negative in the WER models (higher WER) and positive in the logistic exact-transcription models (lower odds of exact transcription). Controls were never removed during model selection.

We fitted the models with \texttt{lme4} (version 2.0.1; \citealt{Bates2015}) in R (version 4.6.0; \citealt{RCoreTeam2025}): logistic \texttt{glmer} models (Laplace approximation, \texttt{bobyqa} optimiser) for \texttt{Exact} and \texttt{lmer} models for WER, with Satterthwaite $p$-values (\texttt{lmerTest}, version 3.2.1; \citealt{Kuznetsova2017}) and 95\% profile intervals; for the logistic models we report log-odds coefficients and
odds ratios with 95\% Wald intervals. The procedure was applied identically to each of the five ASR systems.

\paragraph{Random effects.}
Every model started with a speaker intercept. Where the unit of analysis allowed it, we added the candidate terms of Section~\ref{sec:Experiments} one at a time (intercepts first, then the by-speaker and by-word slopes) and retained a term only if a likelihood-ratio test showed that it improved fit at $\alpha=.05$, the model converged without a negative-eigenvalue warning, and the fit was not singular. Comparisons used maximum likelihood, and the selected \texttt{lmer} models were refitted with REML. 

\paragraph{Additive versus interaction (Experiments~3 and~4).}
The interaction model was retained only if a likelihood-ratio test showed that it improved on the additive model at $\alpha=.05$. Retaining the additive model means that the two effects add on the scale of the model: on the WER scale in Experiment~3, and on the log-odds scale in Experiment~4, where this means that the odds ratios of the two features multiply. A significant interaction means that the features do not combine additively on that scale, and its sign shows whether the combined effect is larger or smaller than expected from the individual effects.

\paragraph{Model criticism.}
We diagnosed the residuals of each selected model individually against pre-specified criteria. The criteria for the linear models are effect sizes and not significance tests, because with thousands of observations, tests of normality have so much power that they reject for trivial departures from normality; for the logistic models a significant test had to be accompanied by a non-trivial effect. For the linear models (standardised conditional residuals) a remedy was needed if at least one criterion was violated: an absolute skewness above 1 \citep{Bulmer1979}; an excess kurtosis above 3; more than 1\% of residuals with an absolute $z$-score above 3; or a ratio above 2 between the residual standard deviations in the highest and lowest quintile of the fitted values. For the logistic models we used simulated scaled residuals from \texttt{DHARMa} \citep{Hartig2022}, and a violation required a significant test ($p<.05$) together with a non-trivial effect (Kolmogorov--Smirnov distance above 0.05, dispersion outside 0.8--1.2, or more than 1\% outliers).
Remedies were applied only when needed, one at a time, and we stopped as soon as the criteria were met: removal of observations with $|z|>3$, then a square-root transformation of WER (for the logistic models, only removal of \texttt{DHARMa} outliers applies). Only three analyses required a remedy, all for \texttt{Whisper tiny.en} in Experiment~1 (1b with all tokens, 1b with single-token utterances, and 1c); in all three, outlier removal did not resolve the violation and the square-root transformation did. The square-root estimates are on a different scale and are compared with the original ones in direction and significance only. All other analyses met the criteria without a remedy.

\paragraph{Final models.}
The final model for each experiment and system, including the models for the copula-type subsets, is given in Table~\ref{tab:final_models} (Appendix~\ref{app:final_models}). The additive model was retained in all main analyses of Experiments~3 and~4 and in all subsets of Experiment~3, for all five systems. The only analysis in which the interaction model was retained is Experiment~4 for wav2vec2 large in the progressive-auxiliary subset.

\FloatBarrier

\section{Results}\label{sec:Results}
\subsection{Overall ASR performance}\label{sec:OverallResult}

Before reporting the results of the feature-specific analyses and the interaction, we first provide an overview of the overall performance of the five ASR systems on our full dataset. We report WER for the target utterances. \texttt{Whisper tiny.en} achieves the lowest WER (28.5\%), followed by wav2vec 2.0 large (42.1\%), wav2vec 2.0 base (47.9\%), Kaldi GigaSpeech L (55.3\%) and Kaldi GigaSpeech S (56.4\%). Within each model family, the system with more capacity or more training data performs better, although the improvement differs in magnitude: wav2vec 2.0 large lowers the WER by 5.8 percentage points relative to wav2vec 2.0 base, whereas Kaldi L improves on Kaldi S by only 1.1 points.

Whisper tiny.en is the smallest system by number of parameters, yet it has the lowest WER, which is consistent with its much larger and more diverse training data. All five systems have WERs well above those reported on their own benchmarks (for example, 16--23\% for the Kaldi systems on GigaSpeech). This indicates that all five systems make substantial errors on our conversational AAE speech, providing a basis for examining whether and how the target linguistic features affect ASR performance in the analyses below.

The following subsections follow the order of the experiments in Section~\ref{sec:StatisticalAnalysis}. Experiments~1 and~2 establish the effect of each feature on its own. Experiments~3 and~4 test the research question, namely whether the co-occurrence of CCR and the null copula increases ASR errors beyond the sum of their individual effects. Under $H_0$, the two effects are additive: the effect of CCR is the same whether the copula is present or absent, so the interaction coefficient is zero. Under $H_1$, the effect of CCR differs with copula status, so the interaction coefficient is non-zero; an interaction that makes errors worse than the sum of the separate effects supports the hypothesis that the two features reinforce each other.

In brief: the null copula raises WER in all five systems (Experiment~2); CCR has a clear effect on whether the CCR token itself is transcribed exactly (Experiments~1a and~4) and a small effect on WER, between 0.6 and 3.2 points when all CCR tokens are analysed and reliable for three of the five systems (Experiment~1b, All), which is not reproduced in the single-token and utterance-level analyses for most systems, wav2vec 2.0 large being the exception; and in neither Experiment~3 nor Experiment~4 did the interaction improve model fit for any system in the full dataset.

\subsection{Experiment 1: CCR alone}\label{sec:ResCCR}

\begin{table*}[!htbp]
\centering
\footnotesize  
\caption{Effect of CCR type in the CCR-only dataset (Experiment~1): odds ratio (canonical vs.\ reduced token) in 1a, WER coefficient in 1b and 1c. $p$-values in parentheses (bold if $<.05$). 1b All: all CCR tokens; 1b Single: utterances with one CCR token; 1c: utterance level, controlling for the number of CCR tokens. $^\dagger$\,Whisper: original models, whose residuals violated our criteria; the square-root models, which met them, gave $p=.018$, $.430$ and $.392$ (1b All, 1b Single, 1c).}
\label{tab:res_ccr}
\begin{tabularx}{\textwidth}{@{}lCCCC@{}}
\toprule
System & 1a: exact token (OR) & 1b All: WER & 1b Single: WER & 1c: utterance WER \\
\midrule
Kaldi L    & 2.16 (\textbf{$<$.001}) & $-$0.006 (.387) & 0.005 (.683) & $-$0.003 (.739) \\
Kaldi S    & 2.14 (\textbf{$<$.001}) & $-$0.013 (.062) & $-$0.008 (.431) & $-$0.008 (.418) \\
w2v2 base  & 1.93 (\textbf{$<$.001}) & $-$0.032 (\textbf{.003}) & $-$0.009 (.443) & $-$0.017 (.082) \\
w2v2 large & 1.56 (\textbf{$<$.001}) & $-$0.014 (\textbf{.049}) & $-$0.024 (\textbf{.033}) & $-$0.029 (\textbf{.002}) \\
Whisper    & 1.23 (\textbf{.006}) & $-$0.027 (\textbf{.011})$^\dagger$ & $-$0.005 (.688)$^\dagger$ & $-$0.005 (.572)$^\dagger$ \\
\bottomrule
\end{tabularx}
\end{table*}

\emph{Experiment 1a.} Canonical CCR tokens were more likely than reduced tokens to be transcribed exactly in all five systems (Table~\ref{tab:res_ccr}). The odds ratio was largest for the two Kaldi systems, smaller for the two wav2vec2 models, with the large model below the base model, and smallest for Whisper, where the effect was nevertheless significant.

\emph{Experiment 1b, all tokens.} The utterance intercept was not retained in any system, because the models did not converge or were singular. The final models therefore contain speaker and word intercepts, plus a by-word slope of CCR type for wav2vec2 base and Whisper. The estimate of CCR type was negative in all five systems, that is, utterances with a reduced token had a higher WER, but its reliability differed (Table~\ref{tab:res_ccr}). It was significant for wav2vec2 base, Whisper and, marginally ($p=.049$), wav2vec2 large, and not for the two Kaldi systems, of which Kaldi S came close ($p=.062$). The effects are small, at most about three WER points. The residual diagnostics met our criteria for every system except Whisper, whose residuals were skewed, heteroscedastic and contained an excess of outliers. For Whisper, the conclusion did not change after removing the 87 outlying residuals, which left the heteroscedasticity unresolved, or under the square-root transformation, which satisfied the criteria (note to Table~\ref{tab:res_ccr}). The significant CCR effects for wav2vec2 base and Whisper in this analysis should be interpreted with caution, for two reasons. First, both come from models with a by-word slope, and their Satterthwaite degrees of freedom are low (89 and 54). Second, each utterance's WER enters the model once for every CCR token it contains, so the standard errors are anti-conservative.

\emph{Experiments 1b single-token and 1c.} Once the repetition is removed, the CCR effects become weaker. In the single-token analysis, CCR type had no significant effect for any system except wav2vec2 large. At the utterance level (1c), wav2vec2 large again showed an effect of reduced CCR, slightly larger than in the single-token analysis, and it remained significant without the control for the number of CCR tokens ($\beta=-0.026$, $p=.004$). No other system showed an effect at the utterance level, although the estimate for wav2vec2 base pointed in the same direction and approached significance. For Whisper, whose residuals violated the criteria in both analyses, outlier removal and the square-root transformation led to the same conclusion as the original model.

In short, reduced CCR lowers the chance that the CCR token itself is transcribed exactly in every system. Its effect on WER is small, at most about three points, and depends on the analysis. When all CCR tokens are analysed, reduced CCR goes with higher WER in all five systems, reliably so in wav2vec2 base and Whisper and, marginally ($p=.049$), in wav2vec2 large. When the repetition is removed, the effects for wav2vec2 base and Whisper disappear. wav2vec2 large is the only system with an effect in all three analyses; the other four show none that survives. The WER effect of CCR is therefore small, and for four of the five systems it disappears once repeated WER values are no longer counted several times.

\subsection{Experiment 2: null copula alone}\label{sec:ResCopula}

As shown in Appendix~\ref{app:final_models}, a by-speaker slope of copula status improved fit and was retained for Kaldi L, wav2vec2 base and wav2vec2 large. It was not retained for Whisper, where it did not improve fit, or for Kaldi S, where the more complex model was singular despite a significant likelihood-ratio test.

Utterances with a null copula had higher WER than utterances with the copula present in all five systems (all $p<.001$; Table~\ref{tab:res_copula}). The difference ranged from 9.7 WER points (Kaldi S) to 13.6 points (Kaldi L). The effect of the null copula on WER was therefore much larger than that of CCR in Experiment~1, where the largest estimate was 3.2 points.

\begin{table*}[!htbp]
\centering
\footnotesize
\caption{Effect of the null copula on utterance WER in the copula-only dataset. Significant coefficients ($p<.05$) are in bold.}
\label{tab:res_copula}
\begin{tabularx}{\textwidth}{@{}l*{5}{C}@{}}
\toprule
System & $\beta$ & 95\% CI & SE & $t$ & $p$ \\
\midrule
Kaldi L & \textbf{$-$0.136} & [$-$0.167, $-$0.106] & 0.015 & $-$8.91 & $<$.001 \\
Kaldi S & \textbf{$-$0.097} & [$-$0.121, $-$0.073] & 0.012 & $-$8.04 & $<$.001 \\
w2v2 base & \textbf{$-$0.111} & [$-$0.142, $-$0.082] & 0.015 & $-$7.31 & $<$.001 \\
w2v2 large & \textbf{$-$0.130} & [$-$0.160, $-$0.103] & 0.014 & $-$9.15 & $<$.001 \\
Whisper & \textbf{$-$0.117} & [$-$0.142, $-$0.094] & 0.012 & $-$9.56 & $<$.001 \\
\bottomrule
\end{tabularx}
\end{table*}

\subsection{Experiment 3: interaction in utterance-level WER}\label{sec:ResInteraction}

As shown in Appendix~\ref{app:final_models}, a by-speaker slope of CCR type improved fit and was retained only for wav2vec2 base and wav2vec2 large. In the additive models, copula status had a significant effect on WER in all five systems (Table~\ref{tab:res_interaction}, upper panel): utterances with a null copula had a higher WER than utterances with an overt copula at both CCR levels (Figure~\ref{fig:interaction}a). CCR type had no significant effect in any system, although its estimate was negative (higher WER for reduced CCR) in every system. For wav2vec2 large, the estimate has the same sign and a similar size to the utterance-level effect of Experiment~1c ($\beta=-0.029$) but is less precise (95\% CI $[-0.048, 0.007]$). The number of CCR tokens in the utterance, included as a control, was not significant in any system (all $p\ge.198$).

Adding the CCR $\times$ copula interaction did not improve fit for any system (Table~\ref{tab:res_interaction}, last column), so the additive model was retained throughout and $H_0$ was not rejected. The interaction coefficients were small and all negative ($-0.010$ to $-0.031$). Because reduced CCR and null copula are both coded $-0.5$, an interaction that makes errors worse than the sum of the two separate effects would appear as a positive coefficient; none of the estimates is positive. Their precision is limited, however: the 95\% confidence intervals lie between $-0.079$ and $+0.037$, so a combined effect up to about eight WER points smaller than the sum cannot be ruled out, nor one up to about four points larger. 

The copula effect in these utterances (3.0 to 9.0 WER points) was smaller than in the copula-only dataset of Experiment~2 (9.7 to 13.6 points). This difference should not be read as evidence about the interaction, because the two datasets differ in composition, for example in utterance length.

\begin{table*}[!htbp]
\centering
\footnotesize
\caption{Additive models and interaction tests in the interaction dataset. Experiment~3: utterance WER (coefficients in proportions of WER; the number of CCR tokens in the utterance is an additional control). Experiment~4: exact transcription of the CCR token (log-odds). $p$-values in parentheses (bold if $<.05$). Interaction: $\chi^2(1)$ of the likelihood-ratio test of the CCR $\times$ NC interaction, with its $p$-value in parentheses.}
\label{tab:res_interaction}
\begin{tabularx}{\textwidth}{@{}llCCC@{}}
\toprule
Experiment & System & CCR type ($\beta$) & NC ($\beta$) & Interaction $\chi^2(1)$ \\
\midrule
\multirow{5}{*}{Experiment~3}
 & Kaldi L & $-$0.005 (.675) & $-$0.055 (\textbf{$<$.001}) & 0.27 (.606) \\
 & Kaldi S & $-$0.012 (.318) & $-$0.050 (\textbf{$<$.001}) & 0.17 (.684) \\
 & w2v2 base & $-$0.022 (.132) & $-$0.030 (\textbf{.037}) & 0.63 (.427) \\
 & w2v2 large & $-$0.020 (.151) & $-$0.051 (\textbf{$<$.001}) & 0.76 (.384) \\
 & Whisper & $-$0.002 (.843) & $-$0.090 (\textbf{$<$.001}) & 1.65 (.199) \\
\midrule
\multirow{5}{*}{Experiment~4}
 & Kaldi L & 0.458 (\textbf{$<$.001}) & 0.292 (\textbf{.004}) & 0.39 (.534) \\
 & Kaldi S & 0.389 (\textbf{$<$.001}) & 0.245 (\textbf{.020}) & 0.50 (.481) \\
 & w2v2 base & 0.584 (\textbf{$<$.001}) & 0.307 (\textbf{.007}) & 1.98 (.160) \\
 & w2v2 large & 0.577 (\textbf{$<$.001}) & 0.360 (\textbf{.002}) & 0.08 (.771) \\
 & Whisper & 0.040 (.700) & 0.472 (\textbf{$<$.001}) & 0.09 (.766) \\
\bottomrule
\end{tabularx}
\end{table*}

\subsection{Experiment 4: interaction at the CCR token}\label{sec:ResToken}

Experiment~4 asks the same question for the exact transcription of the CCR token (Table~\ref{tab:res_interaction}; final models in Appendix~\ref{app:final_models}). Canonical CCR tokens were more likely to be transcribed exactly than reduced tokens in four of the five systems, with the largest coefficients for the two wav2vec2 models and smaller ones for the two Kaldi systems. Whisper showed no CCR effect on the token, although it had one in the CCR-only data of Experiment~1a. Tokens in utterances with a present copula were more likely to be transcribed exactly than tokens in utterances with a null copula in all five systems, and this copula effect was largest in Whisper and smallest in Kaldi S. In the four systems with a CCR effect, the CCR coefficient was larger than the copula coefficient; Whisper was the only system in which the copula, and not CCR, was associated with the transcription of the token. In odds-ratio terms, canonical tokens had about 1.5 to 1.8 times the odds of exact transcription of reduced tokens in the four systems with a CCR effect, and tokens in utterances with a present copula had about 1.3 to 1.6 times the odds of those in utterances with a null copula.

The interaction did not improve fit for any system (Table~\ref{tab:res_interaction}, last column), so the additive model was retained and $H_0$ was not rejected. The interaction test had its lowest $p$-value for wav2vec2 base, but even there it was far from significant. A null copula elsewhere in the utterance is thus associated with a lower probability of exact transcription of the CCR token, and reduced CCR is associated with a lower probability independently of it, with no evidence that either feature strengthens the effect of the other. Figure~\ref{fig:interaction}b shows the predicted probability of an exact transcription of the CCR token, which is higher for canonical than for reduced tokens in four of the five systems (not in Whisper) and higher when the copula is present in all five.

\subsection{Supplementary analyses by copula type}\label{sec:ResING}

The results were the same in nearly all cases when the interaction dataset was split into main-verb copulas and progressive auxiliaries. In Experiment~3 the interaction was not significant in any system in either subset (all $p\ge.164$). In Experiment~4 the interaction was not significant in the main-verb subset (all $p\ge.112$) and in the progressive-auxiliary subset for four systems (all $p\ge.249$). The exception was wav2vec2 large with progressive auxiliaries, where the interaction model fitted significantly better ($\chi^2(1)=4.14$, $p=.042$; interaction $\beta=-0.56$, $SE=0.27$, 95\% CI $[-1.10, -0.03]$). In this subset the advantage of canonical over reduced CCR tokens was larger in utterances with a null copula ($\beta=0.70$) than in utterances with a present copula ($\beta=0.14$), which is the direction predicted by $H_1$ on the log-odds scale.

\begin{figure*}[!t]
\centering
\includegraphics[width=\textwidth]{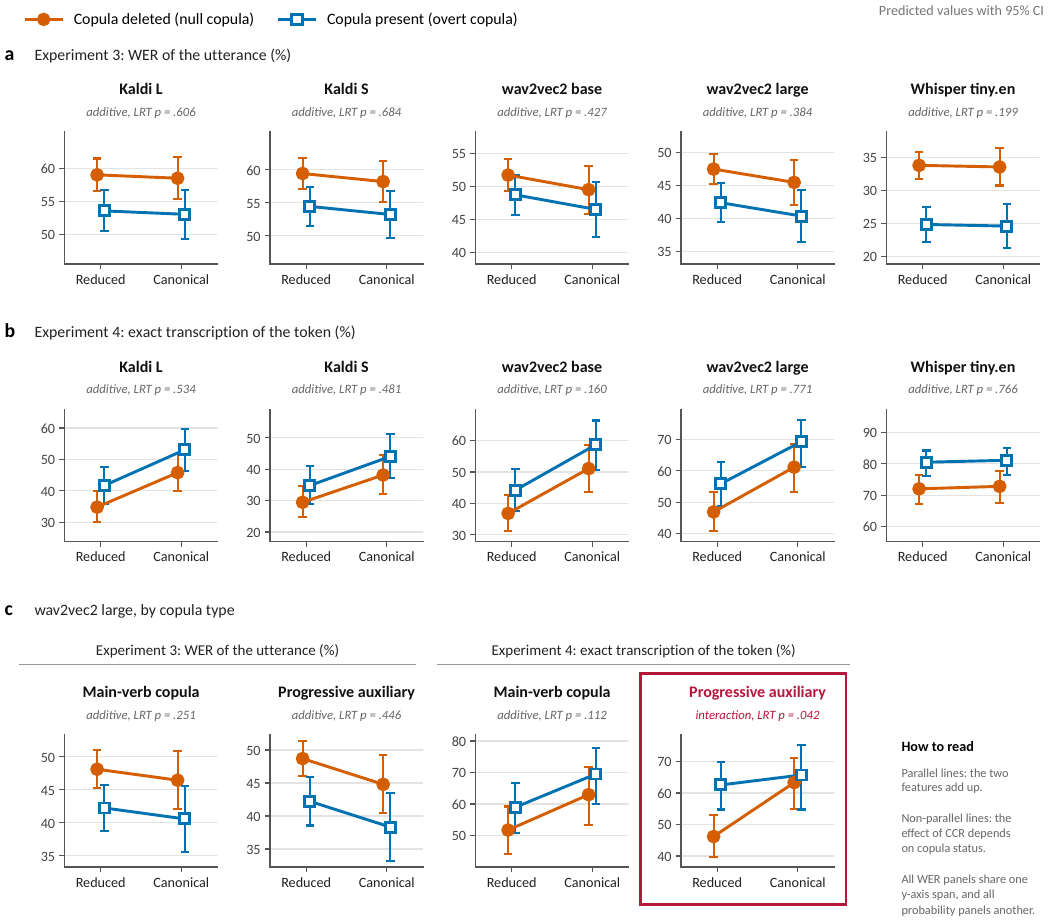}
\caption{Predicted values (\%) with 95\% confidence intervals by CCR type and copula status (open squares: copula present; filled circles: copula deleted). (a)~Experiment~3: WER of the utterance for the five systems. (b)~Experiment~4: exact transcription of the token (probability that the CCR token is transcribed exactly). (c)~wav2vec2 large split by copula type (main-verb copula, progressive auxiliary), for Experiment~3 (left pair) and Experiment~4 (right pair). Each panel shows the model retained by the likelihood-ratio test of the CCR $\times$ copula interaction, with the $p$-value of that test. All models are additive (parallel lines on the model scale) except the framed panel, where the interaction model fitted better. Values are model-based means averaged over age group and gender. The $y$-axes are positioned separately for each panel, but all WER panels share one span and all probability panels share another.}
\label{fig:interaction}
\end{figure*}

\FloatBarrier

This is the framed panel in Figure~\ref{fig:interaction}c: for progressive auxiliaries, reduced and canonical CCR tokens differ little when the copula is present (roughly 3 percentage points) but clearly when it is null (roughly 17 percentage points). The main-verb subset of the same system (Figure~\ref{fig:interaction}c, third panel) does not show this pattern. We treat this result as exploratory. We ran 30 interaction tests (five systems $\times$ three datasets $\times$ two outcomes; the datasets are the full Interaction dataset and the two copula-type subsets). If no interaction existed in any of them, approximately one or two of the 30 tests (at a 5\% false-positive rate per test, $30 \times .05 = 1.5$) would be expected to reach $p<.05$ by chance. No correction for multiple comparisons was applied. Furthermore, the same system did not show a significant interaction in the full Interaction dataset ($p=.771$) or in the main-verb subset ($p=.112$), so this isolated result should not be interpreted as robust evidence for an interaction between CCR and the null copula.

\subsection{Age and gender}\label{sec:ResDemographics}

Age (four groups) and gender were included as controls in every model (final models: Table~\ref{tab:final_models}, Appendix~\ref{app:final_models}); the detailed age and gender coefficients are given in Table~\ref{tab:demographics} in the same appendix. Male speakers had higher WER and lower odds of exact CCR transcription than female speakers in every analysis and system, although the difference was significant in only some of them, most consistently in the interaction dataset (Experiments~3 and~4). Speakers in age groups~2 to~4 generally had lower WER and higher odds of exact CCR transcription than those in age group~1, and every significant age coefficient had this direction, but the pattern varied across systems and experiments, and Whisper showed the fewest significant age effects. Because the age groups are defined separately for each component (e.g., group~2 covers ages 20 to 29 in the Washington, DC components and 30 to 50 in ATL, PRV and ROC; \citealp{farrington_corpus_2021}), a given age group does not correspond to the same ages across components, so age effects partly overlap with differences between components in recording period, locale and conditions; they are best read as adjustments, not as effects of age itself.

\subsection{Summary}\label{sec:ResSummary}

Across the five systems we found no evidence for $H_1$. In utterance-level WER (Experiment~3) and in the exact transcription of the CCR token (Experiment~4), the additive model was preferred in every system, so the two features did not reinforce each other's effects. The null copula was associated with higher WER and lower odds of exact transcription in every system, with a WER effect of about 10 to 14 points in the copula-only dataset (Experiment~2), whereas CCR mainly reduced the exact transcription of the affected token (in four of the five systems in Experiment~4) and had at most a small effect on WER. A single exploratory interaction in one subset of one system does not change this conclusion. Because a non-significant interaction is not proof of additivity, we note that the confidence intervals of the utterance-level interaction estimates (Experiment~3) do not exclude moderate interactions. Age and gender, included as controls, were significantly related to recognition in many analyses (Section~\ref{sec:ResDemographics}), but this does not change the conclusions about the two features.

\section{Discussion}\label{sec:Discussion}

We asked whether the co-occurrence of a phonological feature (CCR) and a morphosyntactic feature (the null copula) increases ASR errors beyond the sum of their individual effects ($H_1$), or whether the two effects simply add up ($H_0$). In our data, they add up. Each feature had an effect of its own (Experiments~1 and~2), but adding the CCR $\times$ copula interaction did not improve model fit for any system, either for utterance WER (Experiment~3) or for exact transcription of the CCR token (Experiment~4), so $H_0$ was not rejected. This conclusion has limits. A non-significant interaction is not proof of additivity: the WER interaction estimates were small and all negative ($-0.010$ to $-0.031$), so none points towards compounding, but their 95\% intervals ($-0.079$ to $+0.037$) do not exclude moderate interactions. Moreover, because CCR hardly moves utterance WER, additive and multiplicative patterns make nearly the same prediction, and Experiment~3 alone could not distinguish them. The token-level analysis is more informative, because CCR lowers exact transcription in four of the five systems, and it likewise showed no interaction. The one exception, wav2vec2 large with progressive auxiliaries ($p=.042$), where the advantage of canonical tokens was larger when the copula was null, is one of 30 interaction tests, was not replicated in the full data or in the main-verb subset, and should be treated as a lead rather than as evidence for $H_1$.

The two features differed greatly in the size and level of their effects on recognition. CCR had its clearest effect on the word that carries it: canonical CCR tokens were more likely than reduced tokens to be transcribed exactly in every system (odds ratios of 1.23 to 2.16, Experiment~1a). Its effect on utterance WER was much smaller and depended on the analysis. By contrast, the null copula raised utterance WER substantially and consistently in every system, by 9.7 to 13.6 points (Experiment~2). When both features occurred in the same utterance, the utterance-level effect was therefore mainly attributable to the null copula. At the token level, both features lowered the odds that the CCR token was transcribed exactly: a reduced CCR token in an utterance with a null copula had lower odds of exact transcription than a canonical token in an utterance with a present copula. However, adding the CCR $\times$ null-copula interaction did not improve model fit in any system in the main analyses, either for utterance WER or for the exact transcription of the CCR token (all $p\ge.160$). Within the precision of our estimates, the effects were therefore approximately additive: the confidence intervals for the interaction estimates include small to moderate interactions, particularly for utterance WER, so additivity here is an approximation and not a demonstrated fact.

Two observations make an additive pattern plausible. First, the features act at different linguistic levels: CCR changes the phonetic form of a word that remains in the utterance, whereas the null copula removes a word and changes utterance structure. Second, their effects differ greatly in size: the smallest null-copula penalty was about three times the largest CCR estimate on utterance WER. At the word that carries it, CCR is a real source of error, but this effect is diluted when the whole utterance is scored. The number of CCR tokens in the utterance, which we controlled for, is only weakly correlated with the CCR status of the utterance ($r=-.24$ in Experiment~3 and $r=-.22$ in Experiment~1c), so collinearity does not explain the small CCR term. Because the two features differ in both their linguistic mechanism and the magnitude of their effects on utterance WER, their co-occurrence is more consistent with additive than interactive effects.

The CCR effect should also be interpreted as an average over heterogeneous cluster types and phonological environments. Reduction is more frequent before consonants than before vowels and in monomorphemic than in bimorphemic clusters, and it is further conditioned by stress, homorganicity, and the type of the following consonant \citep{Erik_Bailey_2015, Wolfram_2017}; the relative weight of these constraints differs between varieties \citep{Wolfram_2017}. With respect to ASR, \citet{mojarad26_interspeech} argue that these factors may differ in their effect on ASR bias. We coded CCR as a binary contrast between canonical and reduced realizations and did not model cluster types or environments, so some of them may have much larger effects on ASR than others, and the small and analysis-dependent CCR effect on utterance WER may partly reflect averaging across such cases. This is a particular concern for the CCR-only dataset, which was constructed to isolate the effect of CCR by excluding utterances containing any copula instance, overt or null; this restriction may have reduced the range of CCR types and phonological environments represented and may have underrepresented those with the largest effects on ASR. It is also a concern for the interaction analyses, which required utterances to contain both CCR and a copula and covered fewer CCR word types.

The contextual-error pattern around habitual \textit{be} reported by \citet{martin20_interspeech} motivated our hypothesis that errors around one feature could propagate to another, but we found no evidence for such propagation at the level tested. We did not measure errors on the words around the feature locations, so local propagation that leaves utterance WER unchanged cannot be excluded. The token-level result does, however, show that the null copula effect extends beyond the position of the copula: in Experiment~4, a null copula elsewhere in the utterance lowered the odds of exact transcription of the CCR token in every system, although neither WER nor the token-level model can identify which words absorb the additional errors. The systems also use context differently (bidirectional rescoring with an explicit $n$-gram and RNN language model in Kaldi, bidirectional attention without a language model in the CTC encoders, and autoregressive generation in Whisper), but our design cannot determine whether these differences explain the absence of an interaction.

From the perspective of ASR architecture, the direction of both effects was the same across the three architecture families, but the magnitude and consistency of their effects differed. Reduced CCR lowered the odds of exact token transcription most strongly in the two Kaldi systems (odds ratios of 2.16 and 2.14), with at most a marginal effect on utterance WER; wav2vec2 large was the only system with a CCR effect on WER in all three analyses; and Whisper showed a CCR effect on WER only in the all-token analysis. Whisper also showed the smallest CCR effect at the token level (odds ratio 1.23 in Experiment~1a; 1.04, not significant, in Experiment~4), a difference that may reflect the different datasets and random-effects structures used in the two experiments and does not show that CCR is irrelevant to Whisper. By contrast, the null copula had a large effect in every system. This pattern is compatible with the view that architecture shapes where a bias appears: \citet{koenecke2020racial} traced racial disparities in commercial ASR mainly to acoustic modelling, \citet{Robertson2024textual} found that wav2vec 2.0 relies more on textual predictability than a hybrid model despite lacking an explicit language model, and \citet{Feng2024} concluded that architecture shapes how bias appears and what mitigation is needed. For wav2vec2 large, the same model class and size as in \citet{mojarad25_interspeech}, our CCR estimates on utterance WER (.014 to .029, depending on the analysis) are of similar size to theirs (.026 to .031). Our results add that a phonological feature can have a clearly architecture-dependent pattern of effects, whereas a morphosyntactic feature that removes a word has a large effect across architectures. 
One reason for the contrast may lie in how much current systems rely on word sequences. \citet{Magaro2026} found that Whisper-large-v3, Parakeet and SeamlessM4T, despite differing architectures, outperformed human listeners on coherent sentences but matched them closely when the surrounding words were scrambled, indicating a shared reliance on statistical dependencies between words. A feature that changes the word sequence, such as the null copula, would then affect all systems, whereas the effect of CCR, which alters a word's acoustic form within an intact sequence, would depend on how well each system recovers the reduced form from context. This fits the weak token-level CCR effect in Whisper and the strong effects in the Kaldi
systems, whose $n$-gram language models capture shorter dependencies. Finally, since the systems also differ in training data and decoding, the differences we observe cannot be attributed to architecture alone.
%Because the systems also differ in training data and decoding, these differences cannot be attributed to architecture alone.

From the perspective of transcription fidelity, part of the CCR effect, especially for wav2vec2, may reflect faithful transcription of non-canonical pronunciations rather than recognition failure. A first, unquantified look at the aligned outputs suggests that wav2vec2 often writes reduced clusters and th-stopping as they were said (e.g., \textit{tes} for \textit{test}, \textit{dat} for \textit{that}), whereas the Kaldi systems and Whisper more often produce a canonical form or a different word. Because CORAAL spells such phonological variants in standard orthography, a faithful transcription is counted as an error. The CCR effect for wav2vec2 may therefore partly reflect this scoring convention. However, the same observation also indicates that wav2vec2 did not consistently recognize reduced forms: it sometimes produced no word or a different word instead. Its CCR effect therefore cannot be interpreted simply as evidence that the system reliably handles CCR, nor can it be attributed entirely to faithful transcription; our analyses do not distinguish between these possibilities. This does not apply to the null copula, which CORAAL transcribes as spoken: there, a faithful transcription is correct, and an added copula is an error.

In terms of the size of the null-copula effect relative to each system's overall WER, the effect was largest in Whisper. It showed the lowest overall WER (28.5\%) and the largest relative null-copula penalty (about 40\% of its mean utterance WER); in absolute terms its penalty in Experiment~2 (11.7 points) was not larger than that of Kaldi~L (13.6) or wav2vec2 large (13.0), but in the interaction dataset it was the largest both in WER points and in log-odds. This pattern is compatible with a baseline effect: systems with high overall WER (55.3\% and 56.4\% for the two Kaldi systems) may absorb an additional error less visibly, whereas Whisper's lower baseline makes an additional copula-related error more apparent. It is also compatible with the possibility that Whisper's decoder substitutes or inserts material at the copula site, as suggested by evidence that its autoregressive decoder can generate material absent from the audio \citep{Koenecke2024careless} and that its output is less verbatim than that of some other systems \citep{heuser24_interspeech}. We did not examine Whisper's output at the position of the copula, however, and because architecture, training data and decoding differ simultaneously, these remain hypotheses rather than tested explanations.

Model size and training data did not behave as one might expect if lower overall WER necessarily reduced feature-specific error effects. Within each model family, the larger or better-trained system improved overall WER, but it did not consistently reduce the effects of CCR or the null copula. For wav2vec2, which shares training data across the two models, the larger model improved overall WER by 5.8 points, yet showed a significant CCR effect on utterance WER in all three CCR-only analyses, whereas the base model showed one only in the repeated-token analysis, which has anti-conservative standard errors. The larger model also had a numerically larger null copula effect in every analysis. For the Kaldi systems, which differ mainly in training data, the larger system likewise improved overall WER but left the CCR effect on the token practically unchanged and produced a numerically larger null copula effect in every analysis. These within-family contrasts do not formally test whether the feature effects differ between systems. They nevertheless show that greater capacity or training data improved overall WER without evidence of a corresponding reduction in the effects of CCR or the null copula. This pattern is consistent with ASR evidence that greater scale or overall accuracy does not necessarily reduce disparities for non-dominant speech: \citet{roll25_slate} found that the relative performance gap for L2 speakers widened as ASR models became larger and more accurate, and \citet{kulkarni24_interspeech} found that Whisper Medium outperformed larger Whisper variants on bias-focused evaluation data. It also agrees with evidence that scale alone does not close the gap for minority varieties \citep{Metzger2026, chang24d_interspeech, vanZee2024group} and that dialect-balanced training data does not guarantee equal performance \citep{lonergan23_interspeech}. A further possible reading is that AAE morphosyntax, which is scarce in the data these systems learn from \citep{dorn-2019-dialect, deas-etal-2025-data, Martin2023}, remains an unfamiliar structure whatever the capacity of the model; we cannot test this with two size contrasts and no control over the training data.

Our results extend earlier feature-level work in three ways. First, they confirm that CCR is a measurable source of ASR error \citep{wassink2022uneven, mojarad25_interspeech}, while showing that its effect is clear at the token that carries it but small at the utterance level. Second, they identify the null copula as a large and consistent error source; to our knowledge, its effect on ASR errors has not previously been estimated in isolation. Its linguistic description may help explain this: \citet{labov_1969} argued that the null copula in AAE parallels contraction in standard English, so systems trained mostly on standard varieties encounter a weak or absent form exactly where they expect \textit{is} or \textit{are}. Third, our additive result is consistent with utterance-level findings that WER rises with the number of AAE features \citep{koenecke2020racial, chang24d_interspeech} and with formalizations of dialect density \citep{johnson22_interspeech}, but it shows that a count of features hides large differences in per-feature effect size. Finally, the two features differ in frequency: CCR is very frequent in AAE, whereas the null copula is less frequent and tends to be more common among speakers from lower social classes \citep{labov_1969}. A small per-token CCR effect can therefore accumulate across many tokens, whereas the larger null copula effect affects fewer utterances. Because our design includes only utterances with at least one feature, it estimates each feature's effect when it occurs and cannot estimate its contribution to total corpus error. The difference in frequency between the two features also means that their effects should be interpreted at different levels: CCR matters because it affects many tokens, whereas the null copula matters because it has a large effect on the utterances in which it occurs.

Taken together, the results support $H_0$ for one phonological and one morphosyntactic feature across three architectures and two size contrasts, but they do not show that all feature combinations behave alike. Two morphosyntactic features, such as the null copula and habitual \textit{be}, could compound because each alters utterance structure; this remains a prediction for future work.  

\subsection{Limitations}\label{sec:limitations}

Our main limitation is that we measured the effects of CCR and the null copula on WER and on the exact transcription of the CCR token, but did not analyse what each system produced at the position of the feature. A full error analysis of how each architecture handles a reduced CCR token and a null copula was beyond the scope of this study. The observations about system outputs in the Discussion, for example that wav2vec2 often writes reduced clusters and th-stopping as they were said, come from a first, unquantified look at the aligned outputs and are not results. Because of this, our CCR effects cannot separate faithful transcription from recognition failure. The same observations suggest, but do not test, that a self-supervised model decoded without a language model could be useful for tagging phonological variation in corpora.

The two features were also measured differently. For CCR, we could score the reduced token itself. The null copula is the absence of a word, so there is no reference token at its position that could be aligned with the ASR output and scored; we therefore measured its effect indirectly, through utterance WER. This mixes errors at the missing-copula position with errors on surrounding words, so we cannot determine whether the effect is local or spread over the utterance. Because WER averages over all words, the size of the copula effect may also depend on utterance length, limiting comparisons across datasets.

The interaction analysis was limited by the rarer feature, the null copula: only copula utterances that also contained CCR could be used, making the Interaction dataset much smaller than the CCR-only dataset. The dataset was also strongly unbalanced with respect to CCR type, with many more reduced than canonical tokens, reducing the information available for estimating the CCR effect and its interaction with copula status, particularly in Experiment~3. These factors limit the precision of the interaction estimates. In addition, the copula contrast is confounded with corpus component and recording conditions \citep{li2024reexamining}, and none of our models includes a component term, so part of the null-copula effect may reflect provenance rather than the feature itself.

Our phonological feature was also narrow: we analysed word-final clusters of two consonants and left three-consonant clusters (e.g., \textit{texts}) for future work. Cluster types (e.g., \textit{st}, \textit{nd}, \textit{ft}) may also differ in how often they reduce and in how they affect ASR output, but we did not analyse them separately, so the pooled CCR effect may hide type-specific effects. The CCR labels came from forced alignment and were checked against human coding only on a small selection of CORAAL (Section~\ref{sec:FeatureExtraction}). Utterances were short, which dilutes the effect of a single word on WER, and they contained other AAE features that we did not annotate.

Finally, our systems covered a limited range: two features in one corpus of interview speech, and five systems of which one was a small Whisper model. We did not include larger Whisper models or LLM-based ASR systems, in which an audio encoder feeds a large language model that generates the transcript (e.g., Qwen3-ASR; \citealp{shi2026qwen3asr}), a quite different architecture. Our conclusions about system differences therefore hold for these five systems only, and the architectural interpretations in the Discussion are hypotheses rather than tested results.

\subsection{Implications and future work}\label{sec:future_work}

For evaluation, assessing each feature separately is informative, because the combined effect of the two features in an utterance is close to the sum of their separate effects. Given that the null copula appears to be the larger source of error in our data, this suggests that, for data collection and system improvement,  morphosyntactic features should be a priority in training and test sets.

Our analysis measures how much each feature raises utterance-level error, but not what the systems do at the feature positions. At a reduced cluster, a system may transcribe the reduced form as spoken, restore the canonical form, substitute another word, or output nothing; at a null copula position, it may restore, omit, or substitute a copula. These outcomes have different consequences for how errors are counted. Restoring the canonical form counts as an error against a verbatim reference but may be the desired output in many applications. Aligning reference and ASR words at these positions would also show whether the effect of a feature is confined to the feature position or spreads to neighbouring words, for example when a system inserts a copula and missegments the words around it. The utterance-level measure cannot separate these cases from errors elsewhere in the utterance.

If restoring canonical forms is driven by a language model's prior over MAE, systems with an external language model should restore more often than systems that decode closer to the acoustic signal. Comparing matched systems with and without an external language model would test this directly. It requires references that accept both variants: scoring phonological variants such as reduced clusters against variant-aware references would separate a system's fidelity to what was said from recognition errors proper. 

On the features and data, three extensions would test how far our conclusions generalise. First, the additivity we found may reflect that CCR and the null copula act at different linguistic levels and rarely interact, and may also reflect their large difference in effect sizes. We should therefore examine the possible interaction between features that occur at the same linguistic level, such as the null copula and person--number disagreement, or CCR and th-stopping. Second, a per-token effect does not by itself rank features by their importance for data collection: a rare feature with a large effect may contribute less to total corpus error than a frequent feature with a small one. Combining each feature's effect with its frequency in running speech would give that ranking directly. Third, we did not model how densely a speaker uses dialect features overall and relied on the speaker-level random effects to capture this variability; future work could instead use the dialect density metric \citep{johnson22_interspeech}.

\section{Conclusion}\label{sec:Conclusion}

This study asked whether the co-occurrence of a phonological and a morphosyntactic feature of African American English increases ASR errors beyond the sum of their individual effects. Across five ASR systems from three architecture families, we found no evidence that consonant cluster reduction and the null copula compounded each other's effects. Their effects were approximately additive, both for utterance WER and for the exact transcription of CCR tokens. Because the confidence intervals of the interaction estimates include small to moderate interactions, these results do not rule out such interactions; the single exploratory interaction found in one subset of one system was not replicated and does not alter this overall conclusion.

The two features nevertheless differed greatly in the size and level of their effects. CCR, a phonological feature, primarily affected the transcription of the word in which it occurred, whereas the null copula, a morphosyntactic deletion, had a large and consistent effect on utterance-level recognition, and a null copula lowered the odds of an exact transcription of a CCR token elsewhere in the utterance in all five systems. This contrast suggests that the consequences of dialectal variation for ASR may depend not only on whether a feature is present, but also on its linguistic level and on whether it alters the form of a retained word or removes a word from the utterance. With one feature of each type, we cannot separate these two possibilities. The larger or better-trained variant within a family (Kaldi L vs.\ S, wav2vec2 large vs.\ base) lowered the overall WER but did not shrink the effect of either feature. Whisper had the lowest overall WER but the largest relative effect of the null copula, which may reflect its lower baseline, although architecture, training data and decoding differ at the same time; such explanations for differences between systems remain hypotheses.

Because the effects were approximately additive, the combined effect of an utterance containing both features can be approximated from their separate effects, and co-occurring dialect features need not be assumed to compound one another. At the same time, the features differed in the size, level, and mechanism of their effects, suggesting that a simple count of minority-variety features may obscure important differences among individual features. On the present evidence, the null copula rather than CCR appears to be the priority for training and test data, and whether this extends to other morphosyntactic features remains to be tested.

Finally, the study highlights the value of evaluating ASR performance at multiple levels. Utterance WER captured the large effect of the null copula but obscured much of the effect of CCR, which was clearest in the exact transcription of the affected token. Feature-level and token-level evaluation can therefore reveal error patterns that aggregate WER does not, and efforts to improve ASR for socially and linguistically diverse speech should evaluate not only overall accuracy, but also how systems handle specific phonological and morphosyntactic variation.

These conclusions are based on two AAE features, five ASR systems, and one corpus of interview speech. Future work should examine additional phonological and morphosyntactic feature combinations, including pairs of features that alter utterance structure, and should determine whether their effects remain additive or interact in other varieties, corpora, and ASR architectures.

\section*{Generative AI Use Disclosure}
No Generative AI tools were used to produce the content or results of this paper. During the preparation of this work the authors used Claude (Anthropic) and Perplexity.ai to improve language and readability, cross-check reported statistics, write and debug code for figures and tables, and to suggest literature. After using these tools, the authors reviewed and edited the content as needed and take full responsibility for the content of the publication.

\appendix

\section{Supplementary material}\label{app:supp}

\setcounter{table}{0}
\renewcommand{\thetable}{A.\arabic{table}}
\renewcommand{\theHtable}{A.\arabic{table}}

\subsection{Final models}\label{app:final_models}

\begingroup
\scriptsize
\setlength{\extrarowheight}{1.5pt}
\setlength{\LTcapwidth}{\textwidth}
\begin{xltabular}{\textwidth}{@{}p{2.3cm}>{\raggedright\arraybackslash}Xp{3.4cm}@{}}

\caption{Final models by experiment and ASR system, including the copula-type subsets (last block). Spkr = speaker; Utt = utterance; CCR\_tok/CCR\_utt = CCR type of the token/utterance; CCR\_total = number of CCR tokens in the utterance; NC = null copula; w2v2 = wav2vec2.}
\label{tab:final_models}\label{tab:final_models_ing}
\\
\toprule
Analysis & Final model & ASR systems \\
\midrule
\endfirsthead
\caption[]{(continued)}\\
\toprule
Analysis & Final model & ASR systems \\
\midrule
\endhead
\midrule
\multicolumn{3}{r}{\emph{continued on next page}}\\
\endfoot
\bottomrule
\endlastfoot
\multicolumn{3}{@{}l}{\textbf{Experiment 1: CCR-only dataset}} \\
\addlinespace[2pt]
1a: exact token
 & \texttt{Exact $\sim$ CCR\_tok + Age + Gender + (1 | Spkr) + (1 + CCR\_tok | Word)}
 & Kaldi L, Kaldi S \\

 & \texttt{Exact $\sim$ CCR\_tok + Age + Gender + (1 | Spkr) + (1 + CCR\_tok | Word) + (1 | Utt)}
 & w2v2 base \\

 & \texttt{Exact $\sim$ CCR\_tok + Age + Gender + (1 | Spkr) + (1 | Word) + (1 | Utt)}
 & w2v2 large \\

 & \texttt{Exact $\sim$ CCR\_tok + Age + Gender + (1 | Spkr) + (1 | Word)}
 & Whisper \\
\addlinespace
1b All tokens: WER
 & \texttt{WER $\sim$ CCR\_tok + Age + Gender + (1 | Spkr) + (1 | Word)}
 & Kaldi L, Kaldi S, w2v2 large \\

 & \texttt{WER $\sim$ CCR\_tok + Age + Gender + (1 | Spkr) + (1 + CCR\_tok | Word)}
 & w2v2 base, Whisper \\
\addlinespace
1b Single-token: WER
 & \texttt{WER $\sim$ CCR\_tok + Age + Gender + (1 | Spkr) + (1 | Word)}
 & Kaldi L, w2v2 base, w2v2 large, Whisper \\

 & \texttt{WER $\sim$ CCR\_tok + Age + Gender + (1 | Spkr)}
 & Kaldi S \\
\addlinespace
1c: Utt WER
 & \texttt{WER $\sim$ CCR\_utt + CCR\_total + Age + Gender + (1 | Spkr)}
 & All five systems \\
\midrule
\multicolumn{3}{@{}l}{\textbf{Experiment 2: Copula-only dataset}} \\
\addlinespace[2pt]
Utt WER
 & \texttt{WER $\sim$ NC + Age + Gender + (1 + NC | Spkr)}
 & Kaldi L, w2v2 base, w2v2 large \\

 & \texttt{WER $\sim$ NC + Age + Gender + (1 | Spkr)}
 & Kaldi S, Whisper \\
\midrule
\multicolumn{3}{@{}l}{\textbf{Experiment 3: Interaction dataset, utterance WER}} \\
\addlinespace[2pt]
Additive model
 & \texttt{WER $\sim$ CCR\_utt + NC + CCR\_total + Age + Gender + (1 | Spkr)}
 & Kaldi L, Kaldi S, Whisper \\

 & \texttt{WER $\sim$ CCR\_utt + NC + CCR\_total + Age + Gender + (1 + CCR\_utt | Spkr)}
 & w2v2 base, w2v2 large \\
\midrule
\multicolumn{3}{@{}l}{\textbf{Experiment 4: Interaction dataset, CCR token}} \\
\addlinespace[2pt]
Additive model
 & \texttt{Exact $\sim$ CCR\_tok + NC + Age + Gender + (1 | Spkr) + (1 | Word)}
 & Kaldi L, Kaldi S \\

 & \texttt{Exact $\sim$ CCR\_tok + NC + Age + Gender + (1 | Spkr) + (1 + CCR\_tok | Word) + (1 | Utt)}
 & w2v2 base, w2v2 large \\

 & \texttt{Exact $\sim$ CCR\_tok + NC + Age + Gender + (1 | Spkr) + (1 | Word) + (1 | Utt)}
 & Whisper \\
\midrule
\multicolumn{3}{@{}l}{\textbf{Copula-type subsets (supplementary)}} \\
\addlinespace[2pt]
Experiment 3: utterance WER, \texttt{ING} = 0 and 1
 & \texttt{Same additive model and random effects as in the full dataset; no interaction retained}
 & All five systems \\
\addlinespace
Experiment 4: main-verb copula (\texttt{ING} = 0)
 & \texttt{Exact $\sim$ CCR\_tok + NC + Age + Gender + (1 | Spkr) + (1 | Word)}
 & Kaldi L, Kaldi S \\

 & \texttt{Exact $\sim$ CCR\_tok + NC + Age + Gender + (1 | Spkr) + (1 | Word) + (1 | Utt)}
 & w2v2 base, Whisper \\

 & \texttt{Exact $\sim$ CCR\_tok + NC + Age + Gender + (1 | Spkr) + (1 + CCR\_tok | Word) + (1 | Utt)}
 & w2v2 large \\
\addlinespace
Experiment 4: progressive auxiliary (\texttt{ING} = 1)
 & \texttt{Exact $\sim$ CCR\_tok + NC + Age + Gender + (1 | Spkr) + (1 | Word)}
 & Kaldi L, Kaldi S, Whisper \\

 & \texttt{Exact $\sim$ CCR\_tok + NC + Age + Gender + (1 | Spkr) + (1 | Word) + (1 | Utt)}
 & w2v2 base \\

 & \texttt{Exact $\sim$ CCR\_tok * NC + Age + Gender + (1 | Spkr) + (1 | Word)}
 & w2v2 large (interaction retained) \\
\end{xltabular}
\endgroup

\subsection{Age and gender coefficients}\label{app:demographics}

\begingroup
\scriptsize
\setlength{\tabcolsep}{3pt}
\renewcommand{\arraystretch}{0.88}
\setlength{\LTcapwidth}{\textwidth}
\begin{xltabular}{\textwidth}{@{}l*{4}{>{\centering\arraybackslash}X}@{}}
\caption{Age and gender coefficients in the final models of Experiments 1--4. Reference levels: age group 1 and female speakers. WER models: proportions of WER; logistic models: log-odds. $p$-values are given in parentheses (bold if $<.05$). $^\dagger$ Whisper: original models (see Table~\ref{tab:res_ccr}).}
\label{tab:demographics}\\
\toprule
System & Age group 2 & Age group 3 & Age group 4 & Male \\
\midrule
\endfirsthead
\caption[]{(continued)}\\
\toprule
System & Age group 2 & Age group 3 & Age group 4 & Male \\
\midrule
\endhead
\midrule
\multicolumn{5}{r}{\emph{continued on next page}}\\
\endfoot
\bottomrule
\endlastfoot
\multicolumn{5}{@{}l}{\textbf{Experiment 1a: Exact CCR transcription (log-odds)}} \\
Kaldi L & 0.554 (\textbf{$<$.001}) & 0.479 (\textbf{.002}) & 0.607 (\textbf{$<$.001}) & $-$0.130 (.274) \\
Kaldi S & 0.637 (\textbf{$<$.001}) & 0.630 (\textbf{$<$.001}) & 0.681 (\textbf{$<$.001}) & $-$0.244 (.055) \\
w2v2 base & 0.448 (\textbf{.015}) & 0.464 (\textbf{.007}) & 0.476 (\textbf{.021}) & $-$0.160 (.238) \\
w2v2 large & 0.650 (\textbf{.004}) & 0.578 (\textbf{.006}) & 0.585 (\textbf{.020}) & $-$0.237 (.150) \\
Whisper & 0.344 (\textbf{.023}) & 0.234 (.099) & 0.129 (.442) & $-$0.227 (\textbf{.041}) \\
\midrule
\multicolumn{5}{@{}l}{\textbf{Experiment 1b: All tokens (WER)}} \\
Kaldi L & $-$0.101 (\textbf{.002}) & $-$0.084 (\textbf{.006}) & $-$0.078 (\textbf{.033}) & 0.059 (\textbf{.014}) \\
Kaldi S & $-$0.106 (\textbf{.001}) & $-$0.095 (\textbf{.002}) & $-$0.078 (\textbf{.031}) & 0.062 (\textbf{.009}) \\
w2v2 base & $-$0.093 (\textbf{.006}) & $-$0.091 (\textbf{.004}) & $-$0.088 (\textbf{.020}) & 0.036 (.143) \\
w2v2 large & $-$0.073 (\textbf{.026}) & $-$0.078 (\textbf{.011}) & $-$0.083 (\textbf{.025}) & 0.051 (\textbf{.033}) \\
Whisper$^\dagger$ & $-$0.056 (\textbf{.020}) & $-$0.038 (.091) & $-$0.024 (.378) & 0.040 (\textbf{.023}) \\
\midrule
\multicolumn{5}{@{}l}{\textbf{Experiment 1b: Single-token utterances (WER)}} \\
Kaldi L & $-$0.108 (\textbf{.001}) & $-$0.085 (\textbf{.007}) & $-$0.062 (.098) & 0.052 (\textbf{.032}) \\
Kaldi S & $-$0.115 (\textbf{$<$.001}) & $-$0.096 (\textbf{.002}) & $-$0.076 (\textbf{.034}) & 0.061 (\textbf{.009}) \\
w2v2 base & $-$0.106 (\textbf{.002}) & $-$0.112 (\textbf{$<$.001}) & $-$0.092 (\textbf{.013}) & 0.035 (.150) \\
w2v2 large & $-$0.086 (\textbf{.007}) & $-$0.088 (\textbf{.004}) & $-$0.074 (\textbf{.040}) & 0.049 (\textbf{.039}) \\
Whisper$^\dagger$ & $-$0.067 (\textbf{.008}) & $-$0.041 (.084) & $-$0.017 (.553) & 0.039 (\textbf{.036}) \\
\midrule
\multicolumn{5}{@{}l}{\textbf{Experiment 1c: Utterance WER}} \\
Kaldi L & $-$0.102 (\textbf{.002}) & $-$0.083 (\textbf{.007}) & $-$0.068 (.061) & 0.055 (\textbf{.021}) \\
Kaldi S & $-$0.107 (\textbf{$<$.001}) & $-$0.095 (\textbf{.002}) & $-$0.076 (\textbf{.034}) & 0.061 (\textbf{.010}) \\
w2v2 base & $-$0.098 (\textbf{.003}) & $-$0.104 (\textbf{$<$.001}) & $-$0.091 (\textbf{.015}) & 0.036 (.140) \\
w2v2 large & $-$0.079 (\textbf{.014}) & $-$0.087 (\textbf{.004}) & $-$0.081 (\textbf{.025}) & 0.049 (\textbf{.038}) \\
Whisper$^\dagger$ & $-$0.059 (\textbf{.015}) & $-$0.041 (.071) & $-$0.019 (.471) & 0.039 (\textbf{.028}) \\
\midrule
\multicolumn{5}{@{}l}{\textbf{Experiment 2: Utterance WER}} \\
Kaldi L & $-$0.047 (.137) & $-$0.071 (\textbf{.018}) & $-$0.057 (.098) & 0.037 (.111) \\
Kaldi S & $-$0.029 (.384) & $-$0.089 (\textbf{.004}) & $-$0.069 (.052) & 0.049 (\textbf{.040}) \\
w2v2 base & 0.014 (.674) & $-$0.059 (.064) & $-$0.037 (.308) & 0.033 (.177) \\
w2v2 large & 0.017 (.584) & $-$0.025 (.385) & $-$0.008 (.823) & 0.016 (.483) \\
Whisper & $-$0.003 (.920) & $-$0.016 (.560) & 0.027 (.383) & 0.025 (.233) \\
\midrule
\multicolumn{5}{@{}l}{\textbf{Experiment 3: Utterance WER}} \\
Kaldi L & $-$0.049 (.126) & $-$0.075 (\textbf{.011}) & $-$0.090 (\textbf{.007}) & 0.059 (\textbf{.011}) \\
Kaldi S & $-$0.046 (.127) & $-$0.058 (\textbf{.040}) & $-$0.060 (.058) & 0.059 (\textbf{.008}) \\
w2v2 base & $-$0.013 (.673) & $-$0.074 (\textbf{.012}) & $-$0.097 (\textbf{.004}) & 0.072 (\textbf{.002}) \\
w2v2 large & $-$0.014 (.636) & $-$0.067 (\textbf{.017}) & $-$0.077 (\textbf{.015}) & 0.066 (\textbf{.002}) \\
Whisper & $-$0.008 (.746) & $-$0.033 (.152) & 0.001 (.976) & 0.042 (\textbf{.021}) \\
\midrule
\multicolumn{5}{@{}l}{\textbf{Experiment 4: Exact CCR transcription (log-odds)}} \\
Kaldi L & 0.374 (\textbf{.027}) & 0.422 (\textbf{.009}) & 0.437 (\textbf{.016}) & $-$0.231 (.061) \\
Kaldi S & 0.175 (.318) & 0.361 (\textbf{.030}) & 0.343 (.067) & $-$0.352 (\textbf{.006}) \\
w2v2 base & 0.112 (.559) & 0.352 (.053) & 0.462 (\textbf{.025}) & $-$0.444 (\textbf{.002}) \\
w2v2 large & 0.388 (\textbf{.049}) & 0.481 (\textbf{.010}) & 0.471 (\textbf{.025}) & $-$0.372 (\textbf{.010}) \\
Whisper & 0.180 (.308) & 0.224 (.185) & $-$0.015 (.938) & $-$0.332 (\textbf{.010}) \\
\end{xltabular}
\endgroup

\printcredits

\bibliographystyle{cas-model2-names}

\bibliography{cas-refs}

\end{document}